\documentclass{article} 
\usepackage[final]{colm2026_conference}

\usepackage{microtype}
\usepackage{hyperref}
\usepackage{url}
\usepackage{booktabs}
\usepackage{siunitx}

\usepackage{lineno}

\usepackage{graphicx}
\usepackage{multicol}
\usepackage{colortbl}
\usepackage{tikz,pgfplots}
\pgfplotsset{compat=newest}
\usepackage{tcolorbox}
\tcbuselibrary{skins}
\usepackage{algorithm}
\usepackage{algpseudocode}
\usepackage{hyperref}
\usepackage{silence}
\usepackage{fontawesome5}
\usepackage{MnSymbol}
\usepackage{multirow, makecell}
\usepackage{hyperref}
\usepackage{amsmath}
\usepackage{algorithm}
\usepackage{algpseudocode}
\usepackage{wrapfig}
\usepackage{pifont}

\definecolor{darkblue}{rgb}{0, 0, 0.5}
\hypersetup{colorlinks=true, citecolor=darkblue, linkcolor=darkblue, urlcolor=darkblue}

\title{Are Large Vision-Language Models Ready to Guide Blind and Low-Vision Individuals?}

\author{
  Eunki Kim$^{1, \dagger}$\thanks{Equal contribution} \quad Na Min An$^{2, }$\footnotemark[1] \quad Wan Ju Kang$^{2}$ \quad Sangryul Kim$^{3, \dagger}$ \\
  \textbf{ James Thorne}$^{4}$ \quad \textbf{Hyunjung Shim}$^{2}$ \\
  $^1$SKT, $^2$KAIST AI, $^3$NAVER, $^4$Theia Insights \\
  \texttt{\{eunkikim, naminan\}@kaist.ac.kr}
}

\usepackage{xspace}
\usepackage{tcolorbox}

\newcommand{\eg}{\emph{e.g.}}
\newcommand{\ie}{\emph{i.e.}}
\newcommand{\oursdata}{\textsc{VL-GuideData}\xspace}
\newcommand{\oursdataM}{\textsc{VL-GuideData-M}\xspace}
\newcommand{\oursdataP}{\textsc{VL-GuideData-P}\xspace}
\newcommand{\oursdataB}{\textsc{VL-GuideData-B}\xspace}
\newcommand{\oursmetric}{\textsc{VL-Guide-S}\xspace}
\newcommand{\oursmetricVLM}{\textsc{VL-Guide-S-VLM}\xspace}

\definecolor{deepblue}{HTML}{6A95B1}
\definecolor{deepred}{HTML}{B16A6A}

\begin{document}

\ifcolmsubmission
\linenumbers
\fi

\maketitle

\begin{abstract}
Large Vision-Language Models (LVLMs) demonstrate a promising direction for assisting individuals with blindness or low-vision (BLV). Yet, measuring their true utility in real-world scenarios is challenging because evaluating whether their descriptions are BLV-informative requires a fundamentally different approach from assessing standard scene descriptions. While the ``VLM-as-a-metric'' or ``LVLM-as-a-judge'' paradigm has emerged, existing evaluators still fall short of capturing the unique requirements of BLV-centric evaluation, lacking at least one of the following key properties: (1) \emph{High correlation with human judgments}, (2) \emph{Long instruction understanding}, (3) \emph{Score generation efficiency}, and (4) \emph{Multi-dimensional assessment}. To this end, we propose a unified framework to bridge the gap between automated evaluation and actual BLV needs. First, we conduct an in-depth user study with BLV participants to understand and quantify their navigational preferences, curating \oursdata, a large-scale BLV user-simulated preference dataset containing image-request-response-score pairs. We then leverage the dataset to develop an accessibility-aware evaluator, \oursmetric, which outperforms existing (L)VLM judges in both human alignment and inference efficiency. Notably, its effectiveness extends beyond a single domain, demonstrating strong performance across multiple fine-grained, BLV-critical dimensions. We hope our work lays as a foundation for automatic AI judges that advance safe, barrier-free navigation for BLV users.
\end{abstract}

\section{Introduction}
``\emph{A good guide is better than a hundred maps.}'' Without reliable assistive guidance, individuals with blindness or low vision (BLV) frequently face unexpected navigational challenges~\citep{brady2013visual,real2019navigation,yuan2024walkvlm,prajapati2024survey}. In complex urban environments, avoiding dynamic obstacles—such as parked vehicles or moving pedestrians—is a matter of physical safety. Surveys indicate that approximately 40\% of legally blind individuals experience accidents or falls at least once a month~\citep{manduchi2011mobility, hogner2015challenges}, with over 90\% of these accidents caused by street obstacles~\citep{wilson2015put, khan2023outdoor}. While AI-based assistive technologies~\citep{be_my_eyes,aira_about_us,seeing_ai,sullivan_pro_app} have increasingly integrated Large Vision-Language Models (LVLMs)~\citep{yuan2024walkvlm, huh2023genassist, zhao2024vialm, liu2025cosight} to provide spatial awareness, a critical gap remains: A standard visual description (\eg, ``A car is parked on the right side of the street.'') may be informative to a sighted user, but it lacks the context-rich, spatially grounded navigation cues that a BLV user requires to actively avoid a collision~\citep{bandukda2019understanding, kazemi2023recognizing}.

\begingroup
\renewcommand\thefootnote{\dagger}
\footnotetext{Work done at KAIST AI.}
\endgroup

Measuring the utility of these models in mobility scenarios is highly challenging because evaluating whether a description is truly \emph{BLV-informative} requires a different modeling perspective than simply assessing a standard scene description. Currently, the automated evaluation of image-text alignment relies on a VLM encoder-based metric or an LVLM judge~\citep{lin2014microsoft, hossain2019comprehensive, ghandi2023deep}. However, existing evaluators fail to meet the unique requirements of BLV-centric evaluation, consistently lacking at least one of four crucial attributes: (1) \emph{High correlation with human judgments}, (2) \emph{Long instruction understanding}, (3) \emph{Score generation efficiency}, and (4) \emph{Multi-dimensional assessment}.

Lightweight encoder-based predictors like CLIP-S~\citep{hessel2021clipscore} and BLIP-S~\citep{li2022blip} are highly efficient but struggle to process the long, detailed text sequences inherent in step-by-step navigational instructions. Conversely, generative LVLMs~\citep{deitke2024molmo, llama3_2} excel at semantic understanding, but their high inference latency and substantial computational requirements limit their practicality for large-scale evaluation. They also often require prompt-tuning and additional post-processing \citep{li2024vlrewardbench, tong2024g, lambert2024rewardbench}, including bias subtraction to prevent ranking distortions \citep{zhu2025charm}. Finally, while recent scalar-based models like InternLM-XComposer-2.5-Reward (IXCREW-S)~\citep{zang2025internlm} improve inference speed, they demonstrate weak pointwise correlations with human judgments on image-text alignment benchmarks~\citep{xu2019structuredmodelingjointdeep, plummer2015flickr30k}. Moreover, the model is tightly coupled with a single LVLM backbone, InternLM, thereby limiting architectural modularity and flexibility.

To address the limitations of existing evaluators that do not fulfill the BLV-centric evaluation criteria, we propose a unified framework that bridges the gap between automated LVLM evaluation and actual BLV needs. Since the unstructured nature of LVLM-generated descriptions can impose significant cognitive overload on BLV users~\citep{chandrasekar2025llm, chen2025surfacing}, we first conduct an in-depth, two-stage user study. By analyzing BLV user preferences across fine-grained dimensions, such as Afraidness and Sufficiency, we isolate exactly what makes an LVLM description useful for safe mobility. We then encapsulate these insights into \oursdata, a novel, large-scale BLV user-simulated preference dataset consisting of 4k image-request-response-score pairs. Unlike existing datasets that evaluate generic image-text alignment~\citep{xu2024visionreward, align_anything} or focus solely on binary Visual Question Answering (VQA)~\citep{yuan2024walkvlm, gurari2019vizwiz}, our \oursdata centers the nuanced perspectives, formatting preferences, and safety criteria of BLV users navigating complex environments.

Additionally, we introduce \oursmetric, a human preference predictor specifically designed to capture BLV navigational utility. \oursmetric satisfies all the above-mentioned four properties required for effective evaluation. It acts as a robust automatic AI judge that rapidly understands long instructions/requests and generates scores consistent with both general sighted judgments and BLV-specific needs. By evaluating across multiple interpretable dimensions rather than providing a single scalar score, \oursmetric outperforms existing (L)VLM judges in human alignment and inference efficiency.

In summary, our contributions are as follows:

\begin{itemize}
    \item Comprehensive analysis of BLV user preferences across diverse LVLM-generated scene descriptions, revealing the fine-grained factors that drive navigational utility.
    
    \item \oursdata, a large-scale multimodal preference dataset curated from BLV-simulated evaluations and specialized mobility scenarios.
    
    \item \oursmetric, an efficient, multi-dimensional LVLM-based predictor that strongly aligns with both sighted and BLV human judgments.
\end{itemize}

\section{Method}
\begin{figure*}[t]
    \centering
    \includegraphics[width=\textwidth]{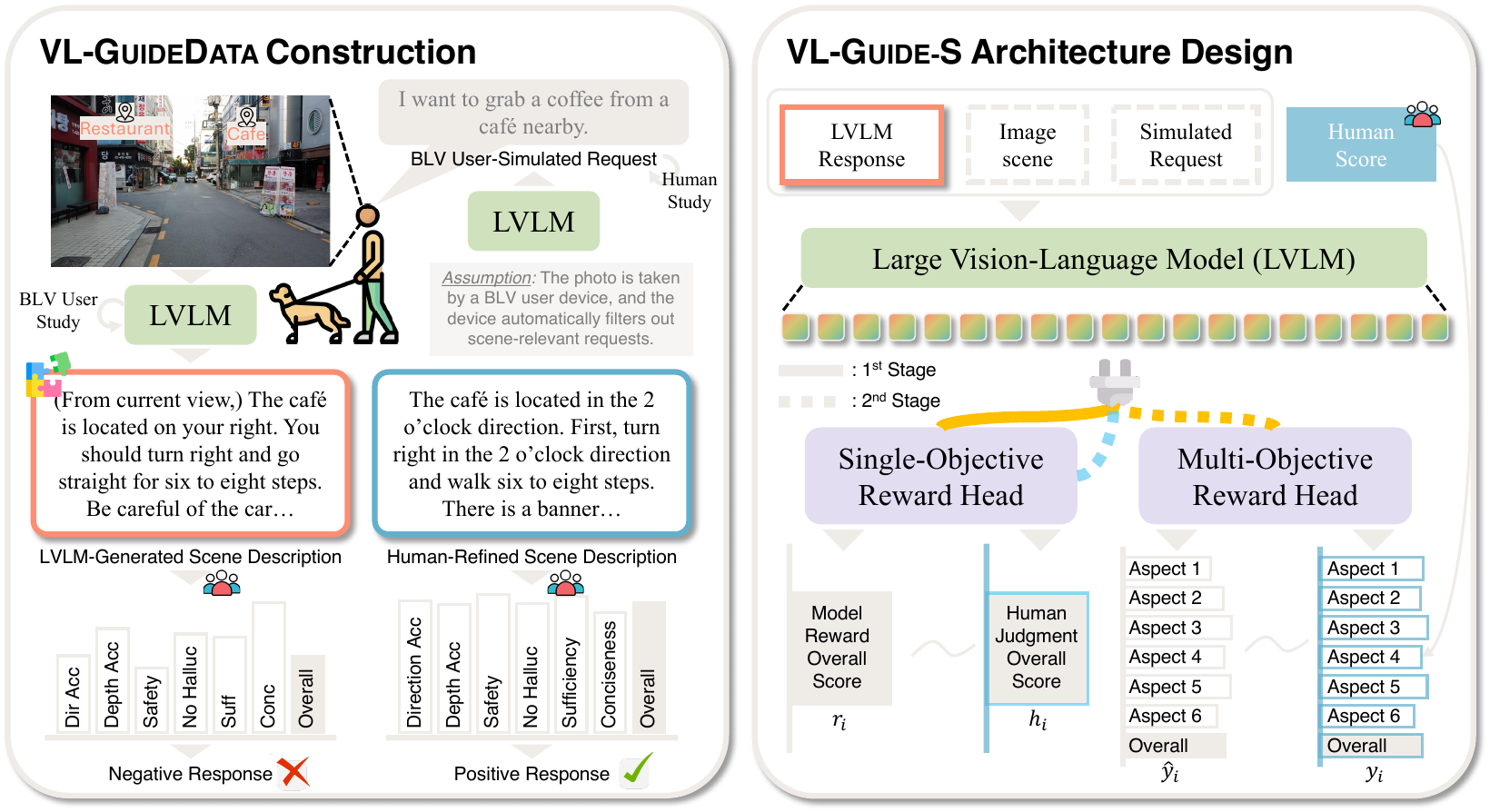}
    \vspace{-1.5em}
    \caption{\textbf{Overall proposed framework.} \textbf{(Left)} Our final benchmark data consist of a scene, a BLV-simulated request, an LVLM-generated/human-refined scene description, and corresponding scores across multiple aspects. \textbf{(Right)} \oursmetric is capable of producing both scalar and multiple values reflecting alignment with human judgments by appending reward heads to a frozen LVLM. The last hidden states of LVLM updated during \emph{stage} 1 are used to train a ridge regression layer during \emph{stage} 2.}
    \vspace{-1em}
    \label{main:framework}
\end{figure*}

To bridge the gap between standard image-text automated evaluation and the specific needs of BLV individuals, we propose a unified framework consisting of two main components (Fig.~\ref{main:framework}): (1) the curation of a BLV user-aware preference dataset (\oursdata) and (2) the development of a BLV user-aligned human preference predictor (\oursmetric).

\subsection{\oursdata: A BLV User-Aware Preference Dataset}

Since collecting real-time, large-scale preference data directly from BLV individuals poses safety concerns and scalability challenges, we adopt a two-phase pipeline. First, we conduct targeted pilot studies with BLV participants to distill their unique navigational requirements into evaluation criteria. We then leverage these insights to guide sighted human annotators in constructing \oursdata, aligned with BLV user-derived guidelines.

\paragraph{(A) Scene Collection and Request Generation.}
We first construct a dataset of environmental scenes and corresponding navigational requests, assuming a realistic assistive setting in which images are captured by a BLV user’s device and automatically paired with scene-relevant requests.
To this end, we filter images from SideGuide~\citep{park2020sideguide} and SideWalk~\citep{aihub_dataset_189}, which contain photos taken by impaired individuals.
We manually curate these datasets to ensure that the images are perceptible to sighted annotators, contain at least five detectable objects, and reflect pedestrian viewpoints (excluding in-vehicle perspectives).
This rigorous four-round filtering process by two authors results in 1,152 high-quality scenes (600 from SideGuide, 552 from SideWalk).

To simulate real-world usage, we generate 5 to 10 plausible BLV navigational requests per scene using GPT-4o mini~\citep{gpt-4omini} with 3-shot prompting. In an Institutional Review Board (IRB)-approved task, 24 sighted annotators review these requests to filter out unnatural or non-actionable prompts (\eg, ``Look in the mirror''). Evaluated by two annotators per pair (62.33\% agreement), this process yields 4,979 consensus image-request pairs (4.32$\pm$1.42 requests per scene). Note that the quality of the automatically generated requests is later also validated by the BLV participants. The choice of having the BLV users verify---rather than directly generate---the requests is deliberate; we focus on how much BLV users \emph{prefer} certain scene descriptions given the identical requests across participants.

\paragraph{(B) Exploratory LVLM Response Generation (\emph{Phase} 1).}
To understand what types of scene descriptions BLV individuals prefer, we initially generate a diverse set of baseline responses. For the image-request pairs, we collect responses from diverse model families, including open-source LVLMs known for strong in-context learning (LLaVA-1.6~\citep{liu2023visualinstructiontuning}, Qwen-VL~\citep{bai2023qwen}, InternLM-XComposer2-VL~\citep{dong2024internlmxcomposer2masteringfreeformtextimage}, and OpenFlamingo~\citep{awadalla2023openflamingoopensourceframeworktraining}) alongside GPT-4o mini~\citep{gpt-4omini}. We apply few-shot prompting on the 7B models and further refine the resulting LVLM responses using GPT-4o mini. This setup enables a controlled evaluation of whether BLV users exhibit a preference for certain model descriptions when the source of generation is not disclosed.

\begin{figure*}[t]
    \centering
    \includegraphics[width=\textwidth]{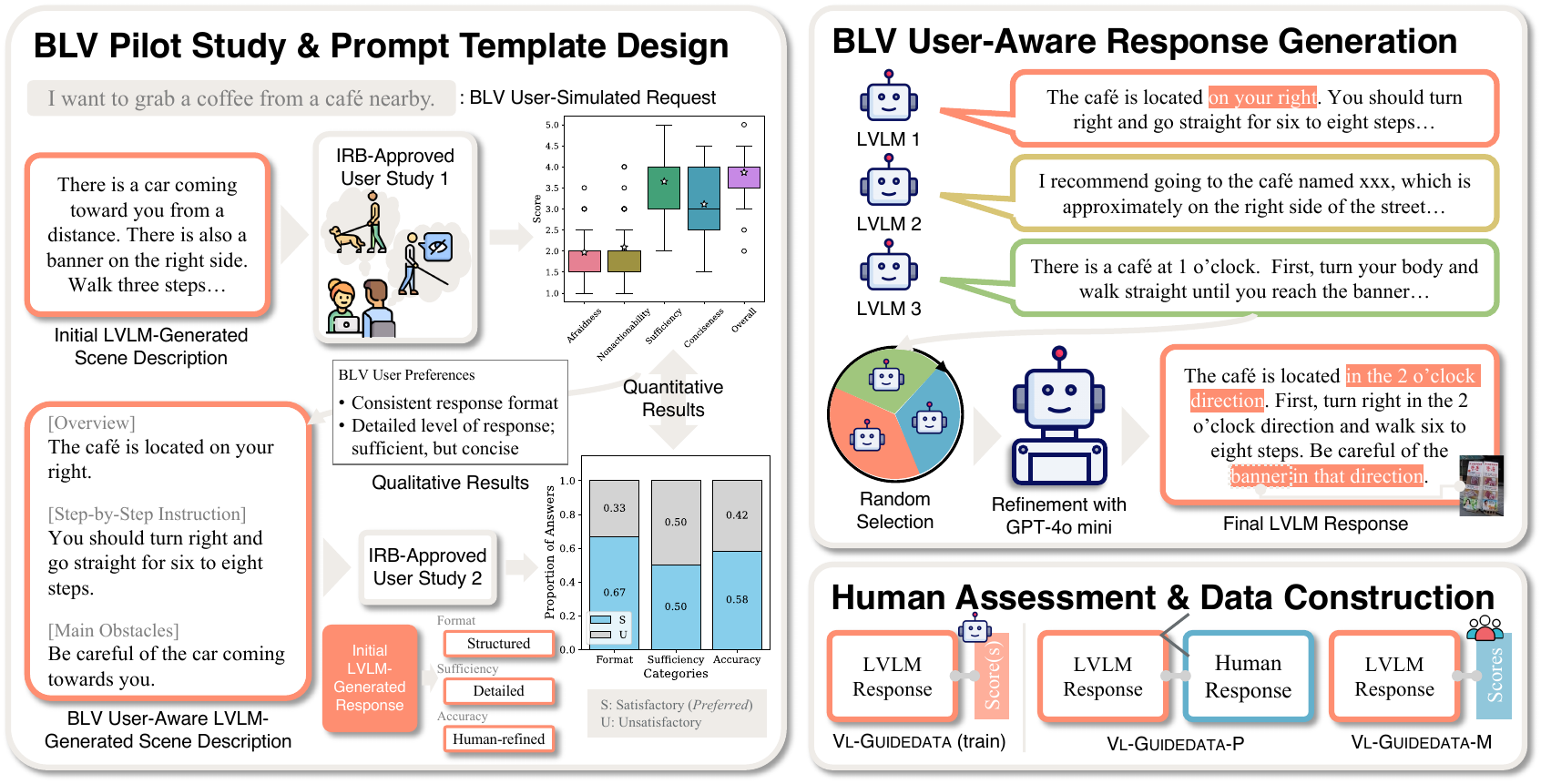}
    \vspace{-1.5em}
    \caption{\textbf{Main results and process of \oursdata construction.} \textbf{(Left)} LVLM prompts are iteratively refined based on BLV user feedback. While raw responses are deemed safe and actionable, users prefer a consistent, succinct format, validated in a second user study. \textbf{(Right)} To reduce model bias, final responses for the \oursdata benchmark are generated by selecting and refining outputs from three different LVLMs. The final LVLM responses are used to construct the training and test set of \oursdata.}
    \vspace{-1.5em}
    \label{main:data_results}
\end{figure*}

\paragraph{(C) BLV User Pilot Study.}

We then present the generated subset of LVLM descriptions to BLV participants across two rounds of IRB-approved user studies. 

In the initial phase, a panel of six BLV participants evaluates the generated scene descriptions across five key metrics: Fear, Actionability, Sufficiency, Conciseness, and Overall Quality. These dimensions are selected to provide a comprehensive assessment, measuring not only the \emph{linguistic quality} of the descriptions (Sufficiency, Conciseness) but also their \emph{psychological and practical} impact on BLV users navigating their environments (Fear, Actionability). The overall results (left panel in Fig.~\ref{main:data_results}) indicate that while the initial LVLM scene descriptions successfully reduce navigational apprehension (Fear $\downarrow$, Actionability $\downarrow$), \emph{how information is conveyed} needs improvement. Specifically, BLV users strongly prefer structured, deductive formats--such as a brief overview followed by specific details--over unstructured text. Furthermore, participants emphasize the need to balance conciseness with adequate detail, prioritizing concrete directional cues and objective distance metrics (\eg, meters or steps) over subjective, generic warnings (\eg, Watch out for cars or pedestrians.) frequently added by GPT-4o mini.

Drawing on initial BLV user feedback, we develop a structured template to systematically reformat LVLM responses (Fig.~\ref{main:data_results}). To determine whether users demonstrate a significant preference for human-edited content, we also incorporate human-refined outputs into our evaluation set. Finally, to validate the hypothesized trade-off between sufficiency and conciseness, we test whether appending specific details to the baseline LVLM responses would yield mixed user preferences. As anticipated, participants strongly prefer template-adherent responses (\emph{Format} in Fig.~\ref{main:data_results}). Surprisingly, they also consistently favor human-edited refinements, even when blinded to the source of the revisions (\emph{Accuracy} in Fig.~\ref{main:data_results}). Consistent with findings from the first round, preferences regarding the balance of detailed versus concise information remained mixed (\emph{Sufficiency} in Fig.~\ref{main:data_results}).

While it is ideal to curate benchmark datasets collected from BLV users, scaling evaluation to a large BLV population is impractical due to recruitment constraints. Hence, we distill the core variables driving BLV user preferences to simulate their conditions and scale our evaluation utilizing a more accessible cohort of sighted participants.

\paragraph{(D) Scaled, BLV user-Aware LVLM Response Generation (\emph{Phase} 2).}

Guided by the key finding from the BLV pilot study, we construct question sets using LVLM to be given to the sighted participants (right panel in Fig.~\ref{main:data_results}). Distinct from the prompts used before the BLV pilot study, we prompt LLaVA-1.6~\citep{liu2023visualinstructiontuning}, Qwen-VL~\citep{bai2023qwen}, and InternLM-XComposer2-VL~\citep{dong2024internlmxcomposer2masteringfreeformtextimage} to adhere to the BLV user-preferred unified template. Then, for each instance, we randomly select one of the three different LVLM responses and refine it by prompting GPT-4o mini~\citep{gpt-4omini}. This process of using GPT-4o to refine responses from different models rather than generating them from scratch helps to minimize the introduction of bias induced by a single model.

\paragraph{(E) Large-Scale Human Assessment (\oursdata Construction).}

To construct \oursdata, we employ 25 sighted human annotators (approved by the Institutional Review Board) to evaluate the refined 2.5k image-request-response triplets from the \emph{Phase} 2 generation. Crucially, since our BLV pilot study reveals that baseline LVLM responses already consistently reduce the fear of BLV users and include BLV user-executable actions, and sighted annotators cannot authentically evaluate subjective BLV experience, we reformulate the pilot metrics into seven visually verifiable dimensions.

Specifically, the BLV requirement for ``Actionability'' is decomposed into (1) \emph{Direction Accuracy} and (2) \emph{Depth Accuracy}, ensuring the spatial guidance is physically correct and executable based on the image.
Similarly, the mitigation of ``Fear'' is enforced through rigorous visual checks for (3) \emph{Safety} (identifying physical hazards) and the absence of (4) \emph{Hallucination} (preventing fabricated, dangerous guidance). The remaining linguistic and structural preferences---(5) \emph{Sufficiency}, (6) \emph{Conciseness}, and (7) \emph{Overall Quality}---are retained directly from the BLV pilot study. All aspects are annotated with an averaged scalar score from two to three human judgments on a Likert scale of 1 to 5, except for \emph{Hallucination}, which is evaluated dichotomously as either 0 (presence) or 1 (absence) based on non-related information, inaccurate step-by-step order, or repeated content.

This robust annotation procedure yields 2,112 unique samples to be scored by 2--3 humans for each.
Additionally, we collect high-quality, human-refined (positive) captions that perfectly satisfy all seven criteria.
After filteration, this procedure yields two evaluation subsets: \oursdataM (LVLM responses labeled with pointwise scores across multiple dimensions; $n=939$) and \oursdataP (pairwise human/LVLM responses; $n=722$). We also generate the training set of \oursdata for \emph{stage} 2 using GPT-4o mini as a scorer for 1.57k image-request-response pairs.

\subsection{\oursmetric: BLV User-Aware Human Judgment-Aligned Evaluator}

Building upon the insights from \oursdata construction, we introduce \oursmetric, an LVLM-based human judgment predictor designed to quantify \emph{the extent to which LVLMs effectively convey scene information for human users, including individuals with blindness or low vision}. Unlike existing reward models that aggregate multiple scores from conflicting objectives into a single score, \oursmetric employs a two-stage architecture for robust single- and multi-objective scoring.

\paragraph{\emph{Stage} 1: Single-Objective Scoring.}
The goal of \emph{stage} 1 is to output a scalar score that captures overall semantic alignment between images and texts, while simultaneously shaping multimodal embeddings useful for the later stage.
As illustrated in Fig.~\ref{main:framework}, we simply append a reward head to a base generative LVLM. This design allows the LVLM to produce a human judgment-aligned multimodal representation in general. Moreover, this architecture inherently supports the comprehension of complex, long-form instructions via the LVLM tokenizer. It also significantly reduces the inference latency compared to text-based generative scoring methods since it eliminating the $\mathcal{O}(k)$ autoregressive decoding overhead required by generative methods to output $k$ tokens, one at a time.

To explicitly align outputs with human judgment, we use Mean Squared Error (MSE) loss:$\min_\theta \sum^N_{i=1} (r_i - h_i)^2$ ($\theta$: param of the LVLM and appended reward head; $h_i$: ground-truth human judgment score; $r_i$: model-predicted scalar score). Initialization on the reward head is conducted using a zero-centered Gaussian distribution with standard deviation $\frac{1}{\sqrt{d+1}}$ ($d$: hidden dimension)~\citep{vonwerra2022trl}. 

\paragraph{\emph{Stage} 2: Multi-Objective Scoring.}
Optimizing evaluators under a multi-objective scoring setting is particularly challenging. Prior work has shown that gradient-based optimization methods (\emph{\eg}, MGDA~\citep{zhang2024mgda}, PCGrad~\citep{yu2020gradient}) are costly at the scale of LVLMs and often fail to balance inherently conflicting objectives~\citep{he2025pareto}. Ensemble-based approaches such as EMORL~\citep{kong2025emorl}, partially alleviate these difficulties by aggregating models trained on individual objectives, but still face instability and limited interpretability without post-hoc calibration when deployed at scale. These observations suggest that relying solely on parameter-level updates or model ensembling is sub-optimal, highlighting the necessity for a lightweight training strategy to effectively scale multi-objective scoring.

To address this, we leverage frozen multimodal embeddings from the \oursmetric architecture trained during \emph{stage} 1 to predict scores across multiple dimensions.
Let $z_i$ denote the frozen multimodal embedding and $y_i$ be the vector of human scores across $K$ dimensions. We append a lightweight ridge regression layer to predict $\hat{y}_i = W z_i + b$, optimized via: $\min_{W,b} \sum_{i=1}^{N} \|y_i - \hat{y}_i\|_2^2 + \alpha\|W\|_F^2$. This formulation enables \oursmetric to produce instantaneous, interpretable scores across diverse criteria, extending beyond a simple image-text alignment score.

Our design of reward modeling introduces a new training paradigm for estimating human judgment scores for multimodal data. Existing methods~\citep{wang2024interpretable,xu2024visionreward}, adopt a \emph{compose-then-aggregate} strategy: they first predict multi-objective scores and then combine them into a single score via a learned linear or gating head. In contrast, we reverse this design by directly supervising a single holistic reward, and subsequently leveraging the model’s final hidden representation to derive scores across multiple objectives.
This design is motivated by the limitation that aggregating predefined dimensions may fail to capture overall quality faithfully and can introduce instability. For example, VisionREW-S exhibits near-zero weights in its aggregation head constantly for certain dimensions, suggesting that some objectives contribute little to the final score. Our approach of projecting the multimodal embedding into dimension-specific scores is inherently more stable and informative, as the trained latent space is already optimized to encode a holistic judgment of humans.

\section{Experiments}

\subsection{Experimental Setup}
To demonstrate architectural agnosticity, we instantiate \oursmetric across several widely used LVLM backbones: Qwen2-VL (2B and 7B), InternLM-XComposer-2.5 (7B), and LLaMA-3.2 (11B). Before adding the layer used for multi-objective scoring, we first train \oursmetric on Polaris~\citep{wada2024polos} and ImgREW~\citep{xu2024imagereward}, which are image-to-text (I2T) and text-to-image (T2I) datasets consisting of human judgment scores. Then, we fine-tune \oursmetric using the training split of our data.

\subsection{Comparison Baselines}
We benchmark \oursmetric against a comprehensive suite of existing evaluators:

\begin{itemize}
    \item \textbf{VLM-based Metrics}: CLIP-S/RefCLIP-S~\citep{hessel2021clipscore}, LongCLIP-S~\citep{zhang2025long}, PAC-S/RefPAC-S~\citep{sarto2024positive}, Polos~\citep{wada2024polos}, BLIP-S~\citep{li2022blip}, and ImgREW-S~\citep{xu2024imagereward}.

    \item \textbf{LVLM-based Generative Judges}: Molmo-7B~\citep{deitke2024molmo}, LLaVA-Critic-7B~\citep{xiong2024llava}, Qwen2-VL-7B~\citep{Qwen2VL}, InternVL2-8B~\citep{chen2024internvl}, and G-VEval~\citep{tong2024g}.
    
    \item \textbf{Scalar Reward Models}: IXCREW-S~\citep{zang2025internlm} for single objective scoring and VisionREW-S~\citep{zhang2024vision} for multiple objective scoring.
\end{itemize}

\subsection{Benchmarks and Metrics}
Evaluations are conducted on standard general-domain datasets (PASCAL~\citep{xu2019structuredmodelingjointdeep}, FOIL~\citep{shekhar-etal-2017-foil}, Flickr-Exp/CF~\citep{plummer2015flickr30k}, Polaris~\citep{wada2024polos}, and ImgREW~\citep{xu2024imagereward}) to ensure baseline competence. To measure true BLV-centric utility, we assess various methods on Sightation~\citep{kang2025sightation} and \oursdata (test). For multi-objective scoring evaluation, we benchmark predictors on VisionREW~\citep{xu2024visionreward}, Align-anything~\citep{align_anything}, and \oursdata.
Metrics used to quantify human-alignment correlation include Pairwise Accuracy (P-Acc) for pairwise ranking datasets and Kendall’s $\tau_b$ or $\tau_c$ for pointwise datasets.


\begin{figure*}[t]
    \centering
    \includegraphics[width=\textwidth]{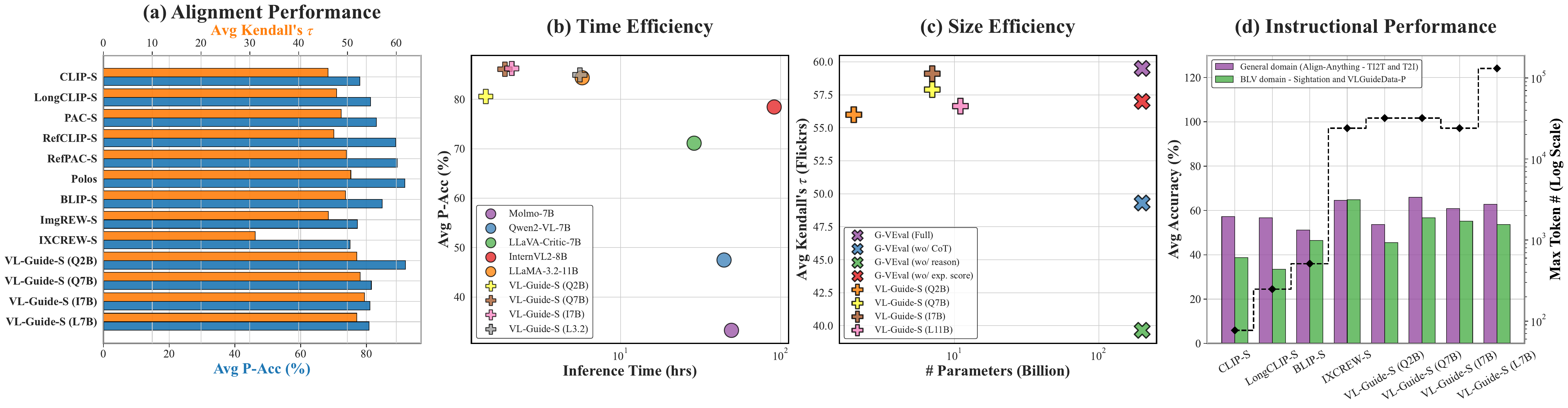}
    \vspace{-2.5em}
    \caption{\textbf{Single-objective performance results.} \textbf{(a)} \oursmetric consistently achieve competitive or superior alignment with human judgments. \textbf{(b)} Ours variants significantly reduce inference time while maintaining high performance. \textbf{(c)} Our open-source \oursmetric achieves comparable human correlation scores to proprietary GPT-4o-based evaluator. \textbf{(d)} \oursmetric is capable of processing long texts while showing high performance in both general and BLV user domains.}
    \vspace{-1.5em}
    \label{main:stage1_results}
\end{figure*}

\subsection{Results}

\subsubsection{Alignment with Human Judgments for a Single Objective}

\paragraph{General image-text datasets.}

\oursmetric significantly outperforms conventional VLM/LVLM-based evaluators and scalar reward models (Fig.~\ref{main:stage1_results}a). For instance, \oursmetric (instantiated with the InternLM-7B) achieves the highest Kendall's rank correlations of 59.3 and 39.5 on FlickrExp and FlickrCF, surpassing IXCREW-S by 28.7\% and 65.1\%, respectively. Furthermore, \oursmetric (Qwen-2B-S) attains near-perfect P-Acc of 98.0\% on FOILR1. These results confirm that our novel architecture yields robust and accurate scores highly aligned with the human judgment across diverse image-text pairs.

Simultaneously, \oursmetric achieves higher efficiency even when compared with the original generative version of LVLM (Fig.~\ref{main:stage1_results}b). For example, while InternVL2-8B takes 90 hours for 14k samples from the Polaris dataset, inference time is greatly reduced by 97.8\% for ours. \oursmetric demonstrates performance on par with the GPT-4o-based judge, G-VEval, while operating without external API costs (Fig.~\ref{main:stage1_results}c).

\paragraph{Instructional image-text datasets.}

Human instructions, particularly navigation requests from BLV users, are inherently complex, often requiring the model to interpret extended, sequential, and multi-step directives. While general image-text datasets typically involve captions that can be evaluated under a single binary variable $y \in {0,1}$ indicating correctness, instructional datasets introduce additional contextual variables $c \sim p(c)$ that influence evaluation criteria. Consequently, the judgment score is better characterized as a conditional distribution $p(y \mid x, c)$ as used in \oursmetric, rather than a marginal $p(y \mid x)$ ($x$: multimodal inputs) used by VLM encoder-based metrics (\eg, CLIP-S) with limited context windows. As shown in Fig.~\ref{main:stage1_results}d, while IXCREW-S achieves slightly higher raw accuracy, this is attributed to its overfitting on preference-matching scores rather than granular instruction following. In contrast, \oursmetric consistently achieves generally high performance in both pointwise and pairwise ranking across general and BLV user-aware domains.

\subsubsection{Alignment with Human Judgments for Multiple Objectives}

When multiple evaluation objectives are considered, the model estimates a joint or factorized distribution over dimensions, \ie, $p(\mathbf{y} \mid x, c)$ or $\prod_{k} p(y_k \mid x, c)$, where each $y_k$ corresponds to a specific evaluation criterion. Although the objectives are not strictly independent, this factorization follows from the chain rule $p(\mathbf{y} \mid x, c) = \prod_{k} p(y_k \mid x, c, y_{<k})$, and approximating the conditional dependencies by $p(y_k \mid x, c, y_{<k}) \approx p(y_k \mid x, c)$ yields the tractable form $\prod_{k} p(y_k \mid x, c)$, which serves as an efficient surrogate to the full joint distribution.

\begin{figure*}[t]
    \centering
    \includegraphics[width=\textwidth]{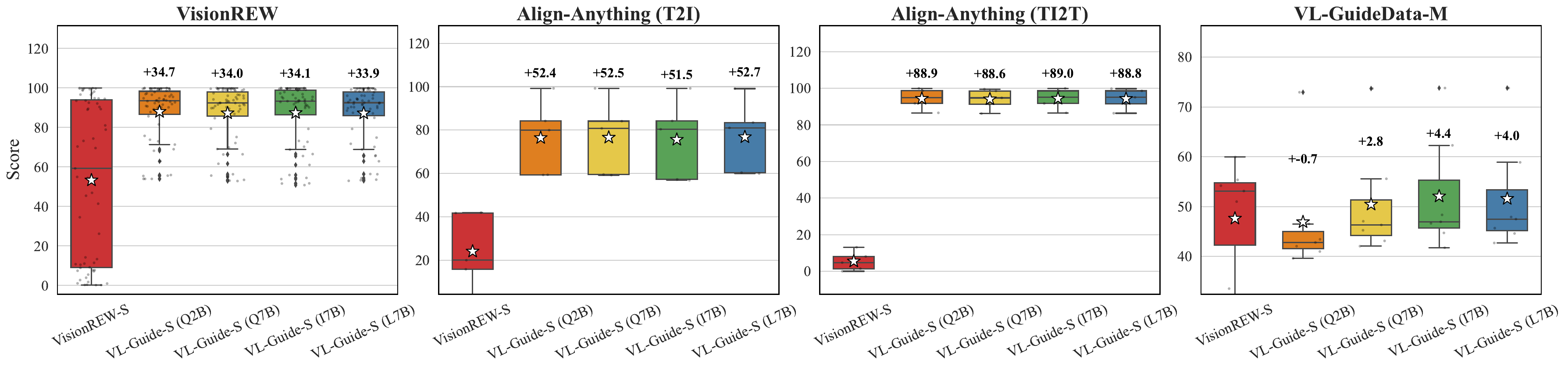}
    \vspace{-2.2em}
    \caption{\textbf{Multi-objective performance results.} The proposed \oursmetric consistently outperforms the baseline across diverse datasets. Furthermore, \oursdataM demonstrates robust scalability as model architecture size increases.}
    \vspace{-1em}
    \label{main:stage2_results}
\end{figure*}

Fig.~\ref{main:stage2_results} shows that \oursmetric outperforms the existing multi-objective reward model, VisionREW-S, in both general and our BLV user-aware preference datasets. 
Similar to the single-objective results, \oursmetric architecture substantially reduces inference time (99.1\%--99.7\%; 51 days \emph{vs}. 4--11 hrs). These results suggest that \oursmetric successfully outputs interpretable, highly accurate assessments across the distinct, fine-grained dimensions necessary for multimodal inputs with additional human/BLV user instructions.

\begin{wraptable}{R}{0.5\textwidth}
\centering
\vspace{-1.2em}
\resizebox{0.5\textwidth}{!}{%
\begin{tabular}{l|c c}
\toprule
Layer Type & \textbf{Acc. ($\uparrow$)} & \textbf{Complexity ($\downarrow$)} \\
\midrule
MLP -- \emph{Baseline} & \textbf{88.6} & \(O(\sum_{l=1}^{L} n_{l-1} n_l)\) \\
Random Forest & 88.2 (\texttt{\textcolor{deepred}{-0.4}}) & \(O(T \cdot D)\) \\
\rowcolor[HTML]{f5f2ee} Ridge Regression & 87.2 (\texttt{\textcolor{deepred}{-1.4}}) & \(\textbf{O(d)}\) \\
\bottomrule
\end{tabular}
}
\vspace{-1em}
\caption{\small{\textbf{The effect of layer types on VisionREW performance (averaged across 59 dimensions) of \oursmetric.} Ridge regression yields the lowest inference complexity while maintaining competitive accuracy (\(d\): input feature dimension; \(L\): \# layers; \(n_l\): \# hidden units in layer \(l\), \(T\): \# trees; \(D\): maximum tree depth.).}}
\vspace{-2em}
\label{tab:ablation}
\end{wraptable}

While we adopt a ridge regression layer to map multimodal embeddings to multi-dimensional scores, this design is not restrictive. As noted in Tab.~\ref{tab:ablation}, replacing it with alternatives such as a Multi-Layer Perceptron (MLP) or a Random Forest regressor yields comparable performance, but introduces additional inference overhead. These results highlight that ridge regression offers the most efficient and scalable choice for multi-dimensional assessment.

\section{Related Work}

\subsection{AI for Blind and Low-Vision Individuals}

AI has increasingly been deployed to assist BLV individuals, with recent advancements in multimodal LVLMs showing potential to transform accessibility~\citep{dong2024internlmxcomposer2masteringfreeformtextimage, zong2025vlicl, liu2024objectfinder}. Applications such as Be My Eyes~\citep{be_my_eyes} and BLV navigation frameworks like WalkVLM~\citep{yuan2024walkvlm} highlight the real-world utility of these barrier-free AI systems. Yet, measuring their true efficacy requires highly representative datasets. Prior datasets, such as VizWiz~\citep{gurari2018vizwiz, tseng2022vizwiz} and BIV-Priv-Seg~\citep{tseng2024biv}, collect images directly from BLV users. While insightful, the reliance on low-resolution or degraded imagery~\citep{bigham2010vizwiz} fails to reflect modern imaging capabilities and artificially restricts the evaluation of advanced AI models~\citep{chiu2020assessing, olson2021towards}. To address this, we focus on high-quality outdoor and indoor scene datasets~\citep{park2020sideguide, aihub_dataset_189} that better represent modern assistive camera inputs.

Furthermore, existing AI systems and datasets predominantly focus on vision-centric tasks, such as standard obstacle detection~\citep{park2020sideguide, xia2023dataset, tang2023dataset} or VQA~\citep{yuan2024walkvlm, gurari2019vizwiz, yang2024viassist}, where outputs are treated as strictly correct or incorrect. These binary formulations do not capture the complexity of generating rich, navigational scene descriptions in dense environments filled with signage and overlapping spatial obstacles. To bridge this gap, we introduce our multimodal BLV preference dataset, \oursdata. Unlike previous datasets, it contains fine-grained BLV navigational preferences alongside qualitative reasoning, shifting the focus from objective scene recognition to subjective, real-world BLV utility. 

\subsection{Automated Evaluators}

Model-based metrics or judges aim to assess image-text alignment by approximating human preferences. Traditional VLM encoder-based metrics are typically divided into reference-based approaches (\eg, Polos~\citep{wada2024polos}, RefPAC-S~\citep{sarto2024positive}), which require costly ground-truth human annotations, and reference-free approaches (\eg, CLIP-S~\citep{hessel2021clipscore}, BLIP-S~\citep{li2022blip}), which score alignment based solely on the image-caption pair. While pointwise and pairwise ranking datasets~\citep{xu2024imagereward, xu2019structuredmodelingjointdeep,tripathi2025pairwise} have enabled nuanced analyses of these metrics, they predominantly focus on general domain alignment rather than specialized user needs. While recent methods like IIT-DAS~\citep{zur2024updating, kreiss2022context} effectively enhances CLIPScore to align with BLV user preferences, it remains limited to the CLIP artchitecture and web-based images. Although CLIP-style architectures are highly efficient, they are inherently not suitable for comprehending the \emph{instructional} texts from the users.

To overcome the limitations of standard evaluators, recent work has shifted toward the ``LVLM-as-a-judge'' paradigm, exploring \emph{generative} LVLMs for evaluation~\citep{deitke2024molmo, xiong2024llava, llama3_2}. For instance,~\citep{tong2024g} utilizes GPT-4o~\citep{gpt-4omini} with Chain-of-Thought (CoT) prompting, while scalar-based reward models have emerged by training linear projection heads on top of LVLMs~\citep{zang2025internlm}. However, existing multi-objective reward models, such as VisionREW-S~\citep{xu2024visionreward} and MPS~\citep{zhang2024learning}, lack in inference efficiency and accessibility. Crucially, because current evaluation datasets lack human judgment scores tailored to the navigational criteria of BLV users~\citep{align_anything, kang2025sightation}, existing LVLM judges fail to perform multi-dimensional assessments that consistently correlate with BLV preferences. We overcome this limitation by introducing \oursmetric, an automated judge designed to evaluate LVLM-generated scene descriptions across multiple dimensions aligned with BLV needs.

\section{Discussion}
In this paper, we introduce a framework for evaluating LVLMs specifically tailored to the needs of BLV users. We demonstrate that grounding evaluation in BLV preferences derived from a series of interviews and human studies faciliates both accurate and efficient assessment. By curating \oursdata and developing \oursmetric, we show that our framework captures nuanced BLV user preferences across single and multiple domains. 

There are several directions in which this work can be extended. First, extending \oursmetric beyond static images to continuous, real-time settings (\eg, evaluating frame-by-frame guidance from egocentric video streams of wearable assistive devices) would better reflect practical deployment scenarios. Second, incorporating user-adaptive evaluation—by explicitly modeling differences in mobility aids (\eg, cane \emph{vs.} guide dog), navigation proficiency, and context (\eg, indoor vs. outdoor or crowded \emph{vs.} sparse environments)—could enable more personalized and inclusive assessment.

With these advancements, we believe that automatic AI judges grounded in real human feedback can more reliably evaluate assistive systems in practice. Ultimately, our goal is to enable vision-language systems that reliably support safe, trustworthy, and barrier-free navigation for all users, including individuals who are blind or have low vision.



\bibliography{colm2026_conference}

@inproceedings{bigham2010vizwiz,
  title={Vizwiz: nearly real-time answers to visual questions},
  author={Bigham, Jeffrey P and Jayant, Chandrika and Ji, Hanjie and Little, Greg and Miller, Andrew and Miller, Robert C and Miller, Robin and Tatarowicz, Aubrey and White, Brandyn and White, Samual and others},
  booktitle={Proceedings of the 23nd annual ACM symposium on User interface software and technology},
  pages={333--342},
  year={2010}
}

@article{zhang2024vision,
  title={Vision-language models for vision tasks: A survey},
  author={Zhang, Jingyi and Huang, Jiaxing and Jin, Sheng and Lu, Shijian},
  journal={IEEE Transactions on Pattern Analysis and Machine Intelligence},
  year={2024},
  publisher={IEEE}
}

@article{Qwen2VL,
  title={Qwen2-VL: Enhancing Vision-Language Model's Perception of the World at Any Resolution},
  author={Wang, Peng and Bai, Shuai and Tan, Sinan and Wang, Shijie and Fan, Zhihao and Bai, Jinze and Chen, Keqin and Liu, Xuejing and Wang, Jialin and Ge, Wenbin and Fan, Yang and Dang, Kai and Du, Mengfei and Ren, Xuancheng and Men, Rui and Liu, Dayiheng and Zhou, Chang and Zhou, Jingren and Lin, Junyang},
  journal={arXiv preprint arXiv:2409.12191},
  year={2024}
}

@article{xiong2024llava,
  title={Llava-critic: Learning to evaluate multimodal models},
  author={Xiong, Tianyi and Wang, Xiyao and Guo, Dong and Ye, Qinghao and Fan, Haoqi and Gu, Quanquan and Huang, Heng and Li, Chunyuan},
  journal={arXiv preprint arXiv:2410.02712},
  year={2024}
}

@article{deitke2024molmo,
  title={Molmo and pixmo: Open weights and open data for state-of-the-art multimodal models},
  author={Deitke, Matt and Clark, Christopher and Lee, Sangho and Tripathi, Rohun and Yang, Yue and Park, Jae Sung and Salehi, Mohammadreza and Muennighoff, Niklas and Lo, Kyle and Soldaini, Luca and others},
  journal={arXiv preprint arXiv:2409.17146},
  year={2024}
}

@article{zang2025internlm,
  title={InternLM-XComposer2. 5-Reward: A Simple Yet Effective Multi-Modal Reward Model},
  author={Zang, Yuhang and Dong, Xiaoyi and Zhang, Pan and Cao, Yuhang and Liu, Ziyu and Ding, Shengyuan and Wu, Shenxi and Ma, Yubo and Duan, Haodong and Zhang, Wenwei and others},
  journal={arXiv preprint arXiv:2501.12368},
  year={2025}
}

@article{xu2024visionreward,
  title={Visionreward: Fine-grained multi-dimensional human preference learning for image and video generation},
  author={Xu, Jiazheng and Huang, Yu and Cheng, Jiale and Yang, Yuanming and Xu, Jiajun and Wang, Yuan and Duan, Wenbo and Yang, Shen and Jin, Qunlin and Li, Shurun and others},
  journal={arXiv preprint arXiv:2412.21059},
  year={2024}
}

@inproceedings{zhang2024learning,
  title={Learning multi-dimensional human preference for text-to-image generation},
  author={Zhang, Sixian and Wang, Bohan and Wu, Junqiang and Li, Yan and Gao, Tingting and Zhang, Di and Wang, Zhongyuan},
  booktitle={Proceedings of the IEEE/CVF Conference on Computer Vision and Pattern Recognition},
  pages={8018--8027},
  year={2024}
}

@inproceedings{olson2021towards,
  title={Towards Using Live Photos to Mitigate Image Quality Issues In VQA Photography},
  author={Olson, Lauren and Kambhamettu, Chandra and McCoy, Kathleen},
  booktitle={Proceedings of the 23rd International ACM SIGACCESS Conference on Computers and Accessibility},
  pages={1--3},
  year={2021}
}

@inproceedings{huh2023genassist,
  title={GenAssist: Making image generation accessible},
  author={Huh, Mina and Peng, Yi-Hao and Pavel, Amy},
  booktitle={Proceedings of the 36th Annual ACM Symposium on User Interface Software and Technology},
  pages={1--17},
  year={2023}
}

@article{zhao2024vialm,
  title={Vialm: A survey and benchmark of visually impaired assistance with large models},
  author={Zhao, Yi and Zhang, Yilin and Xiang, Rong and Li, Jing and Li, Hillming},
  journal={arXiv preprint arXiv:2402.01735},
  year={2024}
}

@article{tripathi2025pairwise,
  title={Pairwise or pointwise? evaluating feedback protocols for bias in llm-based evaluation},
  author={Tripathi, Tuhina and Wadhwa, Manya and Durrett, Greg and Niekum, Scott},
  journal={arXiv preprint arXiv:2504.14716},
  year={2025}
}

@article{liu2024objectfinder,
  title={ObjectFinder: Open-Vocabulary Assistive System for Interactive Object Search by Blind People},
  author={Liu, Ruiping and Zhang, Jiaming and Sch{\"o}n, Angela and M{\"u}ller, Karin and Zheng, Junwei and Yang, Kailun and Gerling, Kathrin and Stiefelhagen, Rainer},
  journal={arXiv preprint arXiv:2412.03118},
  year={2024}
}

@misc{seeing_ai,
  author       = {Microsoft},
  title        = {Seeing AI - Talking Camera App for the Blind},
  year         = {2017},
  url          = {https://www.seeingai.com/}
}

@misc{aira_about_us,
  author       = {Aira},
  title        = {About Us and Our Values},
  year         = {2015},
  url          = {https://aira.io/aira-about-us/},
}

@misc{sullivan_pro_app,
  author       = {TUAT},
  title        = {Sullivan A},
  year         = {2024},
  url          = {https://play.google.com/store/apps/details?id=tuat.kr.sullivan.pro&hl=ko}
}

@inproceedings{bandukda2019understanding,
  title={Understanding experiences of blind individuals in outdoor nature},
  author={Bandukda, Maryam and Singh, Aneesha and Berthouze, Nadia and Holloway, Catherine},
  booktitle={Extended Abstracts of the 2019 CHI Conference on Human Factors in Computing Systems},
  pages={1--6},
  year={2019}
}

@article{kazemi2023recognizing,
  title={Recognizing the Viewpoint and Experience of Blind People in Navigation and Daily Traffic},
  author={Kazemi, Homa and Kamali, Mohammad and Salehi, Reza and Mobaraki, Hossein},
  journal={Function and Disability Journal},
  volume={6},
  number={1},
  pages={0--0},
  year={2023},
  publisher={Function and Disability Journal}
}

@inproceedings{lin2014microsoft,
  title={Microsoft coco: Common objects in context},
  author={Lin, Tsung-Yi and Maire, Michael and Belongie, Serge and Hays, James and Perona, Pietro and Ramanan, Deva and Doll{\'a}r, Piotr and Zitnick, C Lawrence},
  booktitle={Computer Vision--ECCV 2014: 13th European Conference, Zurich, Switzerland, September 6-12, 2014, Proceedings, Part V 13},
  pages={740--755},
  year={2014},
  organization={Springer}
}

@inproceedings{plummer2015flickr30k,
  title={Flickr30k entities: Collecting region-to-phrase correspondences for richer image-to-sentence models},
  author={Plummer, Bryan A and Wang, Liwei and Cervantes, Chris M and Caicedo, Juan C and Hockenmaier, Julia and Lazebnik, Svetlana},
  booktitle={Proceedings of the IEEE international conference on computer vision},
  pages={2641--2649},
  year={2015}
}

@misc{Label,
  title={{Label Studio}: Data labeling software},
  url={https://github.com/heartexlabs/label-studio},
  note={Open source software available from https://github.com/heartexlabs/label-studio},
  author={
    Maxim Tkachenko and
    Mikhail Malyuk and
    Andrey Holmanyuk and
    Nikolai Liubimov},
  year={2020-2022},
}

@article{real2019navigation,
  title={Navigation systems for the blind and visually impaired: Past work, challenges, and open problems},
  author={Real, Santiago and Araujo, Alvaro},
  journal={Sensors},
  volume={19},
  number={15},
  pages={3404},
  year={2019},
  publisher={MDPI}
}

@inproceedings{brady2013visual,
  title={Visual challenges in the everyday lives of blind people},
  author={Brady, Erin and Morris, Meredith Ringel and Zhong, Yu and White, Samuel and Bigham, Jeffrey P},
  booktitle={Proceedings of the SIGCHI conference on human factors in computing systems},
  pages={2117--2126},
  year={2013}
}

@inproceedings{
zong2025vlicl,
title={{VL}-{ICL} Bench: The Devil in the Details of Multimodal In-Context Learning},
author={Yongshuo Zong and Ondrej Bohdal and Timothy Hospedales},
booktitle={The Thirteenth International Conference on Learning Representations},
year={2025},
url={https://openreview.net/forum?id=cpGPPLLYYx}
}

@article{wang2024interpretable,
  title={Interpretable Preferences via Multi-Objective Reward Modeling and Mixture-of-Experts},
  author={Wang, Haoxiang and Xiong, Wei and Xie, Tengyang and Zhao, Han and Zhang, Tong},
  journal={arXiv preprint arXiv:2406.12845},
  year={2024}
}

@article{kang2025sightation,
  title={Sightation Counts: Leveraging Sighted User Feedback in Building a BLV-aligned Dataset of Diagram Descriptions},
  author={Kang, Wan Ju and Kim, Eunki and An, Na Min and Kim, Sangryul and Choi, Haemin and Kwak, Ki Hoon and Thorne, James},
  journal={arXiv preprint arXiv:2503.13369},
  year={2025}
}

@article{li2024vlrewardbench,
  title={VLRewardBench: A Challenging Benchmark for Vision-Language Generative Reward Models},
  author={Li, Lei and Wei, Yuancheng and Xie, Zhihui and Yang, Xuqing and Song, Yifan and Wang, Peiyi and An, Chenxin and Liu, Tianyu and Li, Sujian and Lin, Bill Yuchen and others},
  journal={arXiv preprint arXiv:2411.17451},
  year={2024}
}

@article{lambert2024rewardbench,
  title={Rewardbench: Evaluating reward models for language modeling},
  author={Lambert, Nathan and Pyatkin, Valentina and Morrison, Jacob and Miranda, LJ and Lin, Bill Yuchen and Chandu, Khyathi and Dziri, Nouha and Kumar, Sachin and Zick, Tom and Choi, Yejin and others},
  journal={arXiv preprint arXiv:2403.13787},
  year={2024}
}

@misc{align_anything,
  author = {PKU-Alignment Team},
  title = {Align Anything: training all modality models to follow instructions with unified language feedback},
  year = {2024},
  publisher = {GitHub},
  journal = {GitHub repository},
  howpublished = {\url{https://github.com/PKU-Alignment/align-anything}},
}

@article{hossain2019comprehensive,
  title={A comprehensive survey of deep learning for image captioning},
  author={Hossain, MD Zakir and Sohel, Ferdous and Shiratuddin, Mohd Fairuz and Laga, Hamid},
  journal={ACM Computing Surveys (CsUR)},
  volume={51},
  number={6},
  pages={1--36},
  year={2019},
  publisher={ACM New York, NY, USA}
}

@article{zhu2025charm,
  title={CHARM: Calibrating Reward Models With Chatbot Arena Scores},
  author={Zhu, Xiao and Tan, Chenmien and Chen, Pinzhen and Sennrich, Rico and Zhang, Yanlin and Hu, Hanxu},
  journal={arXiv preprint arXiv:2504.10045},
  year={2025}
}

@inproceedings{chen2024internvl,
  title={Internvl: Scaling up vision foundation models and aligning for generic visual-linguistic tasks},
  author={Chen, Zhe and Wu, Jiannan and Wang, Wenhai and Su, Weijie and Chen, Guo and Xing, Sen and Zhong, Muyan and Zhang, Qinglong and Zhu, Xizhou and Lu, Lewei and others},
  booktitle={Proceedings of the IEEE/CVF Conference on Computer Vision and Pattern Recognition},
  pages={24185--24198},
  year={2024}
}

@misc{zhang2025long,
  title={Long-clip: Unlocking the long-text capability of clip},
  author={Zhang, Beichen and Zhang, Pan and Dong, Xiaoyi and Zang, Yuhang and Wang, Jiaqi},
  booktitle={European Conference on Computer Vision},
  pages={310--325},
  year={2025},
  organization={Springer}
}

@misc{kreiss2022context,
  title={Context Matters for Image Descriptions for Accessibility: Challenges for Referenceless Evaluation Metrics},
  author={Kreiss, Elisa and Bennett, Cynthia and Hooshmand, Shayan and Zelikman, Eric and Morris, Meredith Ringel and Potts, Christopher},
  booktitle={Proceedings of the 2022 Conference on Empirical Methods in Natural Language Processing},
  pages={4685--4697},
  year={2022}
}

@misc{wilson2015put,
  title={Who Put That There: The barrier to blind and partially sighted people getting out and about},
  author={Wilson, M},
  journal={RNIB: London, UK},
  year={2015}
}

@misc{liu2023visualinstructiontuning,
      title={Visual Instruction Tuning}, 
      author={Haotian Liu and Chunyuan Li and Qingyang Wu and Yong Jae Lee},
      year={2023},
      eprint={2304.08485},
      archivePrefix={arXiv},
      primaryClass={cs.CV},
      url={https://arxiv.org/abs/2304.08485}, 
}

@misc{wang2024qwen2vlenhancingvisionlanguagemodels,
      title={Qwen2-VL: Enhancing Vision-Language Model's Perception of the World at Any Resolution}, 
      author={Peng Wang and Shuai Bai and Sinan Tan and Shijie Wang and Zhihao Fan and Jinze Bai and Keqin Chen and Xuejing Liu and Jialin Wang and Wenbin Ge and Yang Fan and Kai Dang and Mengfei Du and Xuancheng Ren and Rui Men and Dayiheng Liu and Chang Zhou and Jingren Zhou and Junyang Lin},
      year={2024},
      eprint={2409.12191},
      archivePrefix={arXiv},
      primaryClass={cs.CV},
      url={https://arxiv.org/abs/2409.12191}, 
}

@misc{awadalla2023openflamingoopensourceframeworktraining,
      title={OpenFlamingo: An Open-Source Framework for Training Large Autoregressive Vision-Language Models}, 
      author={Anas Awadalla and Irena Gao and Josh Gardner and Jack Hessel and Yusuf Hanafy and Wanrong Zhu and Kalyani Marathe and Yonatan Bitton and Samir Gadre and Shiori Sagawa and Jenia Jitsev and Simon Kornblith and Pang Wei Koh and Gabriel Ilharco and Mitchell Wortsman and Ludwig Schmidt},
      year={2023},
      eprint={2308.01390},
      archivePrefix={arXiv},
      primaryClass={cs.CV},
      url={https://arxiv.org/abs/2308.01390}, 
}

@misc{zur2024updating,
  title={Updating CLIP to Prefer Descriptions Over Captions},
  author={Zur, Amir and Kreiss, Elisa and D'Oosterlinck, Karel and Potts, Christopher and Geiger, Atticus},
  journal={arXiv preprint arXiv:2406.09458},
  year={2024}
}

@inproceedings{shekhar-etal-2017-foil,
    title = "{FOIL} it! Find One mismatch between Image and Language caption",
    author = "Shekhar, Ravi  and
      Pezzelle, Sandro  and
      Klimovich, Yauhen  and
      Herbelot, Aur{\'e}lie  and
      Nabi, Moin  and
      Sangineto, Enver  and
      Bernardi, Raffaella",
    editor = "Barzilay, Regina  and
      Kan, Min-Yen",
    booktitle = "Proceedings of the 55th Annual Meeting of the Association for Computational Linguistics (Volume 1: Long Papers)",
    month = jul,
    year = "2017",
    address = "Vancouver, Canada",
    publisher = "Association for Computational Linguistics",
    url = "https://aclanthology.org/P17-1024/",
    doi = "10.18653/v1/P17-1024",
    pages = "255--265",
}

@misc{chen2025surfacing,
  title={Surfacing Variations to Calibrate Perceived Reliability of MLLM-generated Image Descriptions},
  author={Chen, Meng and Iyer, Akhil and Pavel, Amy},
  journal={arXiv preprint arXiv:2507.15692},
  year={2025}
}

@misc{chandrasekar2025llm,
  title={LLM impact on BLV programming},
  author={Chandrasekar, Prashant and Couvillion, Mariel and Saktheeswaran, Ayshwarya and Zeitz, Jessica},
  journal={arXiv preprint arXiv:2504.17018},
  year={2025}
}

@misc{liu2025cosight,
  title={CoSight: Exploring Viewer Contributions to Online Video Accessibility Through Descriptive Commenting},
  author={Liu, Xingyu and Wang, Biao and Zhang, Wayne and Liao, Ziqian and Li, Ziwen and Pavel, Amy and Chen, Xiang'Anthony' and others},
  journal={arXiv preprint arXiv:2508.08582},
  year={2025}
}

@article{manduchi2011mobility,
  title={Mobility-related accidents experienced by people with visual impairment},
  author={Manduchi, Roberto and Kurniawan, Sri},
  journal={AER Journal: Research and Practice in Visual Impairment and Blindness},
  volume={4},
  number={2},
  pages={44--54},
  year={2011}
}

@article{hogner2015challenges,
  title={Challenges in traffic for blind and visually impaired people and strategies for their safe participation},
  author={H{\"o}gner, N},
  journal={Klinische Monatsblatter fur Augenheilkunde},
  volume={232},
  number={8},
  pages={982--987},
  year={2015}
}

@article{khan2023outdoor,
  title={Outdoor mobility aid for people with visual impairment: Obstacle detection and responsive framework for the scene perception during the outdoor mobility of people with visual impairment},
  author={Khan, Wasiq and Hussain, Abir and Khan, Bilal Muhammad and Crockett, Keeley},
  journal={Expert Systems with Applications},
  volume={228},
  pages={120464},
  year={2023},
  publisher={Elsevier}
}

@misc{bai2023qwen,
  title={Qwen technical report},
  author={Bai, Jinze and Bai, Shuai and Chu, Yunfei and Cui, Zeyu and Dang, Kai and Deng, Xiaodong and Fan, Yang and Ge, Wenbin and Han, Yu and Huang, Fei and others},
  journal={arXiv preprint arXiv:2309.16609},
  year={2023}
}

@inproceedings{chiu2020assessing,
  title={Assessing image quality issues for real-world problems. In 2020 IEEE},
  author={Chiu, Tai-Yin and Zhao, Yinan and Gurari, Danna},
  booktitle={CVF Conference on Computer Vision and Pattern Recognition (CVPR)},
  pages={3643--3653},
  year={2020}
}

@article{tseng2024biv,
  title={BIV-Priv-Seg: Locating Private Content in Images Taken by People With Visual Impairments},
  author={Tseng, Yu-Yun and Sharma, Tanusree and Zhang, Lotus and Stangl, Abigale and Findlater, Leah and Wang, Yang and Gurari, Danna},
  journal={arXiv preprint arXiv:2407.18243},
  year={2024}
}

@inproceedings{gurari2019vizwiz,
  title={Vizwiz-priv: A dataset for recognizing the presence and purpose of private visual information in images taken by blind people},
  author={Gurari, Danna and Li, Qing and Lin, Chi and Zhao, Yinan and Guo, Anhong and Stangl, Abigale and Bigham, Jeffrey P},
  booktitle={Proceedings of the IEEE/CVF Conference on Computer Vision and Pattern Recognition},
  pages={939--948},
  year={2019}
}

@inproceedings{tseng2022vizwiz,
  title={Vizwiz-fewshot: Locating objects in images taken by people with visual impairments},
  author={Tseng, Yu-Yun and Bell, Alexander and Gurari, Danna},
  booktitle={European Conference on Computer Vision},
  pages={575--591},
  year={2022},
  organization={Springer}
}

@inproceedings{gurari2018vizwiz,
  title={Vizwiz grand challenge: Answering visual questions from blind people},
  author={Gurari, Danna and Li, Qing and Stangl, Abigale J and Guo, Anhong and Lin, Chi and Grauman, Kristen and Luo, Jiebo and Bigham, Jeffrey P},
  booktitle={Proceedings of the IEEE conference on computer vision and pattern recognition},
  pages={3608--3617},
  year={2018}
}

@inproceedings{li2022blip,
  title={Blip: Bootstrapping language-image pre-training for unified vision-language understanding and generation},
  author={Li, Junnan and Li, Dongxu and Xiong, Caiming and Hoi, Steven},
  booktitle={International conference on machine learning},
  pages={12888--12900},
  year={2022},
  organization={PMLR}
}

@inproceedings{wada2024polos,
  title={Polos: Multimodal Metric Learning from Human Feedback for Image Captioning},
  author={Wada, Yuiga and Kaneda, Kanta and Saito, Daichi and Sugiura, Komei},
  booktitle={Proceedings of the IEEE/CVF Conference on Computer Vision and Pattern Recognition},
  pages={13559--13568},
  year={2024}
}

@article{sarto2024positive,
  title={Positive-Augmented Contrastive Learning for Vision-and-Language Evaluation and Training},
  author={Sarto, Sara and Moratelli, Nicholas and Cornia, Marcella and Baraldi, Lorenzo and Cucchiara, Rita},
  journal={arXiv preprint arXiv:2410.07336},
  year={2024}
}

@article{xu2024imagereward,
  title={Imagereward: Learning and evaluating human preferences for text-to-image generation},
  author={Xu, Jiazheng and Liu, Xiao and Wu, Yuchen and Tong, Yuxuan and Li, Qinkai and Ding, Ming and Tang, Jie and Dong, Yuxiao},
  journal={Advances in Neural Information Processing Systems},
  volume={36},
  year={2024}
}

@inproceedings{hessel2021clipscore,
  title={CLIPScore: A Reference-free Evaluation Metric for Image Captioning},
  author={Hessel, Jack and Holtzman, Ari and Forbes, Maxwell and Le Bras, Ronan and Choi, Yejin},
  booktitle={Proceedings of the 2021 Conference on Empirical Methods in Natural Language Processing},
  pages={7514--7528},
  year={2021}
}

@article{yang2024viassist,
  title={VIAssist: Adapting Multi-modal Large Language Models for Users with Visual Impairments},
  author={Yang, Bufang and He, Lixing and Liu, Kaiwei and Yan, Zhenyu},
  journal={arXiv preprint arXiv:2404.02508},
  year={2024}
}

@article{tang2023dataset,
  title={A dataset for the recognition of obstacles on blind sidewalk},
  author={Tang, Wu and Liu, De-er and Zhao, Xiaoli and Chen, Zenghui and Zhao, Chen},
  journal={Universal Access in the Information Society},
  volume={22},
  number={1},
  pages={69--82},
  year={2023},
  publisher={Springer}
}

@article{xia2023dataset,
  title={A dataset for the visually impaired walk on the road},
  author={Xia, Haiying and Yao, Cong and Tan, Yumei and Song, Shuxiang},
  journal={Displays},
  volume={79},
  pages={102486},
  year={2023},
  publisher={Elsevier}
}

@misc{be_my_eyes,
  author       = {Wiberg, Hans Jørgen},
  title        = {Be My Eyes - See the World Together},
  year         = {2015},
  url          = {https://www.bemyeyes.com/}
}

@misc{aihub_dataset_189,
  author       = {AIHub},
  title        = {AI Hub: Sidewalk Dataset},
  year         = {2019},
  url          = {https://aihub.or.kr/aihubdata/data/view.do?currMenu=115&topMenu=100&aihubDataSe=realm&dataSetSn=189}
}

@inproceedings{park2020sideguide,
  title={Sideguide: a large-scale sidewalk dataset for guiding impaired people},
  author={Park, Kibaek and Oh, Youngtaek and Ham, Soomin and Joo, Kyungdon and Kim, Hyokyoung and Kum, Hyoyoung and Kweon, In So},
  booktitle={2020 IEEE/RSJ International Conference on Intelligent Robots and Systems (IROS)},
  pages={10022--10029},
  year={2020},
  organization={IEEE}
}

@misc{dong2024internlmxcomposer2masteringfreeformtextimage,
      title={InternLM-XComposer2: Mastering Free-form Text-Image Composition and Comprehension in Vision-Language Large Model}, 
      author={Xiaoyi Dong and Pan Zhang and Yuhang Zang and Yuhang Cao and Bin Wang and Linke Ouyang and Xilin Wei and Songyang Zhang and Haodong Duan and Maosong Cao and Wenwei Zhang and Yining Li and Hang Yan and Yang Gao and Xinyue Zhang and Wei Li and Jingwen Li and Kai Chen and Conghui He and Xingcheng Zhang and Yu Qiao and Dahua Lin and Jiaqi Wang},
      year={2024},
      eprint={2401.16420},
      archivePrefix={arXiv},
      primaryClass={cs.CV},
      url={https://arxiv.org/abs/2401.16420}, 
}

@misc{gpt-4omini,
title={GPT-4o mini},
author={OpenAI},
url={https://openai.com/index/gpt-4o-mini-advancing-cost-efficient-intelligence/},
year={2024b},
}

@inproceedings{prajapati2024survey,
  title={A Survey on Navigation Assistance System for Visually Impaired and Blind People},
  author={Prajapati, Devanshi and Bordoloi, Prapti and Sheth, Rushil and Sharma, Ankit K},
  booktitle={International Conference on Information and Communication Technology for Intelligent Systems},
  pages={569--577},
  year={2024},
  organization={Springer}
}

@article{yuan2024walkvlm,
  title={WalkVLM: Aid Visually Impaired People Walking by Vision Language Model},
  author={Yuan, Zhiqiang and Zhang, Ting and Zhang, Jiapei and Zhou, Jie and Zhang, Jinchao},
  journal={arXiv preprint arXiv:2412.20903},
  year={2024}
}

@misc{llama3_2,
    title = {Llama 3.2},
    author = {Meta},
    url = {https://ai.meta.com/blog/llama-3-2-connect-2024-vision-edge-mobile-devices/},
    year={2024}
}

@misc{xu2019structuredmodelingjointdeep,
      title={Structured Modeling of Joint Deep Feature and Prediction Refinement for Salient Object Detection}, 
      author={Yingyue Xu and Dan Xu and Xiaopeng Hong and Wanli Ouyang and Rongrong Ji and Min Xu and Guoying Zhao},
      year={2019},
      eprint={1909.04366},
      archivePrefix={arXiv},
      primaryClass={cs.CV},
      url={https://arxiv.org/abs/1909.04366}, 
}

@article{tong2024g,
  title={G-VEval: A Versatile Metric for Evaluating Image and Video Captions Using GPT-4o},
  author={Tong, Tony Cheng and He, Sirui and Shao, Zhiwen and Yeung, Dit-Yan},
  journal={arXiv preprint arXiv:2412.13647},
  year={2024}
}

@article{ghandi2023deep,
  title={Deep learning approaches on image captioning: A review},
  author={Ghandi, Taraneh and Pourreza, Hamidreza and Mahyar, Hamidreza},
  journal={ACM Computing Surveys},
  volume={56},
  number={3},
  pages={1--39},
  year={2023},
  publisher={ACM New York, NY}
}

@misc{vonwerra2022trl,
  author = {Leandro von Werra and Younes Belkada and Lewis Tunstall and Edward Beeching and Tristan Thrush and Nathan Lambert and Shengyi Huang and Kashif Rasul and Quentin Gallouédec},
  title = {TRL: Transformer Reinforcement Learning},
  year = {2020},
  publisher = {GitHub},
  journal = {GitHub repository},
  howpublished = {\url{https://github.com/huggingface/trl}}
}

@article{he2025pareto,
  title={Pareto Multi-Objective Alignment for Language Models},
  author={He, Qiang and Maghsudi, Setareh},
  journal={arXiv preprint arXiv:2508.07768},
  year={2025}
}

@article{kong2025emorl,
  title={EMORL: Ensemble Multi-Objective Reinforcement Learning for Efficient and Flexible LLM Fine-Tuning},
  author={Kong, Lingxiao and Yang, Cong and Neufang, Susanne and Beyan, Oya Deniz and Boukhers, Zeyd},
  journal={arXiv preprint arXiv:2505.02579},
  year={2025}
}

@article{zhang2024mgda,
  title={MGDA converges under generalized smoothness, provably},
  author={Zhang, Qi and Xiao, Peiyao and Zou, Shaofeng and Ji, Kaiyi},
  journal={arXiv preprint arXiv:2405.19440},
  year={2024}
}

@article{yu2020gradient,
  title={Gradient surgery for multi-task learning},
  author={Yu, Tianhe and Kumar, Saurabh and Gupta, Abhishek and Levine, Sergey and Hausman, Karol and Finn, Chelsea},
  journal={Advances in neural information processing systems},
  volume={33},
  pages={5824--5836},
  year={2020}
}
\bibliographystyle{colm2026_conference}

\newpage
\appendix

\renewcommand{\thefigure}{S\arabic{figure}}
\renewcommand{\thetable}{S\arabic{table}}
\renewcommand{\theequation}{S\arabic{equation}}
\setcounter{figure}{0}
\setcounter{table}{0}
\setcounter{equation}{0}

\section*{Appendix}

Due to the limited pages, we provide supplementary materials in the Appendix with the following contents:

\begin{itemize}
  \item Section~\ref{app:bench}: Dataset Construction Details
  \item Section~\ref{app:model}: Method Implementation Details
  \item Section~\ref{app:res}: Additional Results
  \item Section~\ref{app:lim}: Discussion
\end{itemize}


\section{Dataset Construction Details}\label{app:bench}

Our proposed benchmark dataset is initially curated by filtering images from open-source BLV-simulated sidewalk views~\citep{park2020sideguide, aihub_dataset_189}. These images are then paired with BLV-simulated mobility requests, LVLM-generated scene descriptions, direct BLV user preferences (\oursdataB), and BLV-simulated human preferences (\oursdataM; \oursdataP). Below, we detail each stage of this dataset construction process. All human studies are conducted under protocols approved by the IRB.

\subsection{BLV-Simulated Requests}\label{app:requests}

\paragraph{Model generation.}
We outline the details of constructing our final 4,979 mobility request datasets. First, we prompt GPT-4o mini (\$0.15/1M input tokens)~\citep{gpt-4omini} using the prompt in Tab.~\ref{app:system_prompts_scengen}. The model is provided with 3-shot examples for both outdoor (Tab.~\ref{app:fewshot_prompts_scengen_outdoors}) and indoor (Tab.~\ref{app:fewshot_prompts_scengen_indoors}) environments. We set the temperature to 0.0 and the maximum tokens to 300. This initial stage generates 8,149 requests for 1,150 images, yielding 7.09 requests per image. These raw requests are subsequently filtered by 24 human annotators, resulting in the final 4,979 image-request pairs.

\paragraph{Human study.}

\begin{figure*}[ht]
    \centering
    \includegraphics[width=\textwidth]{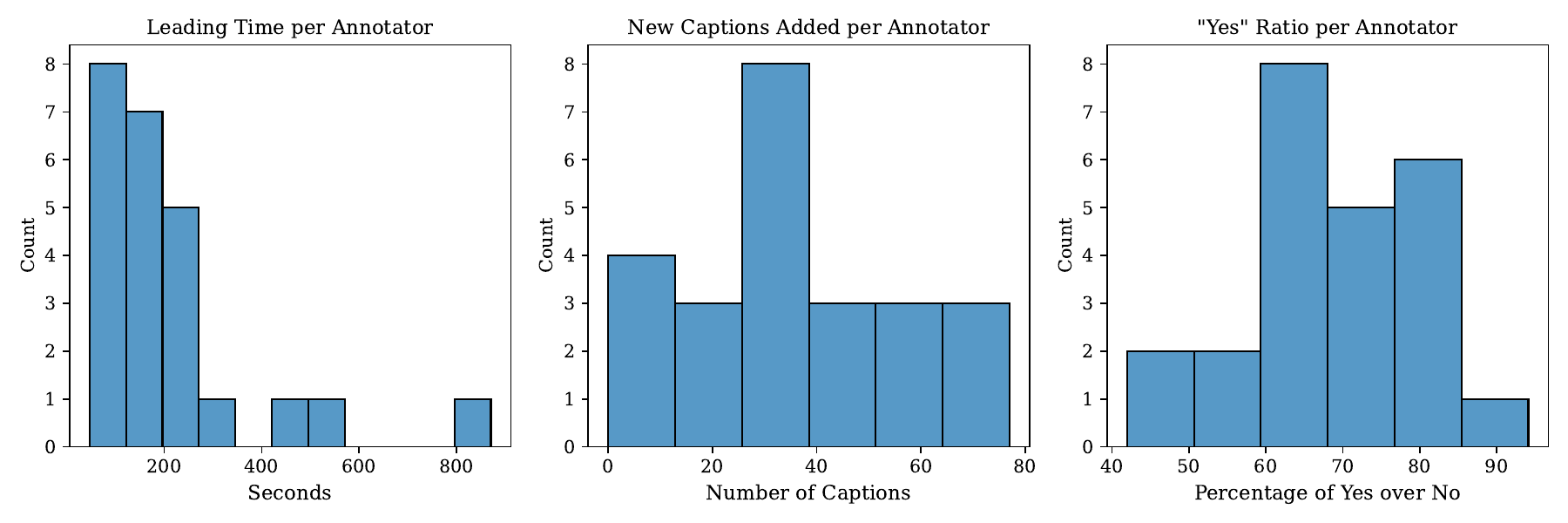}
    \vspace{-2em}
    \caption{\textbf{Distribution of human annotation statistics.} The histograms illustrate annotator behavior across three key metrics: (left) the annotation time spent per sample, showing most tasks are completed within 300 seconds; (middle) the total number of new request captions added by each annotator, highlighting varying levels of proactive contribution; and (right) the acceptance rate (``Yes'' ratio) for verifying simulated requests, indicating that annotators typically accepted 60\% to 80\% of the generated samples.}
    \label{app:human_dist}
\end{figure*}

To rigorously filter the GPT-4o mini outputs and verify their visual relevance to the corresponding images, we recruit 24 sighted human annotators. Annotators are provided with detailed instructions (Tab.~\ref{app:request_guideline}) and are compensated with 50,000 KRW ($\sim$34 USD as of January 2025). The task duration ranges from under an hour to two hours per participant (\emph{see} leading times in Fig.~\ref{app:human_dist}).

\begin{figure*}[ht]
    \centering
    \includegraphics[width=0.7\textwidth]{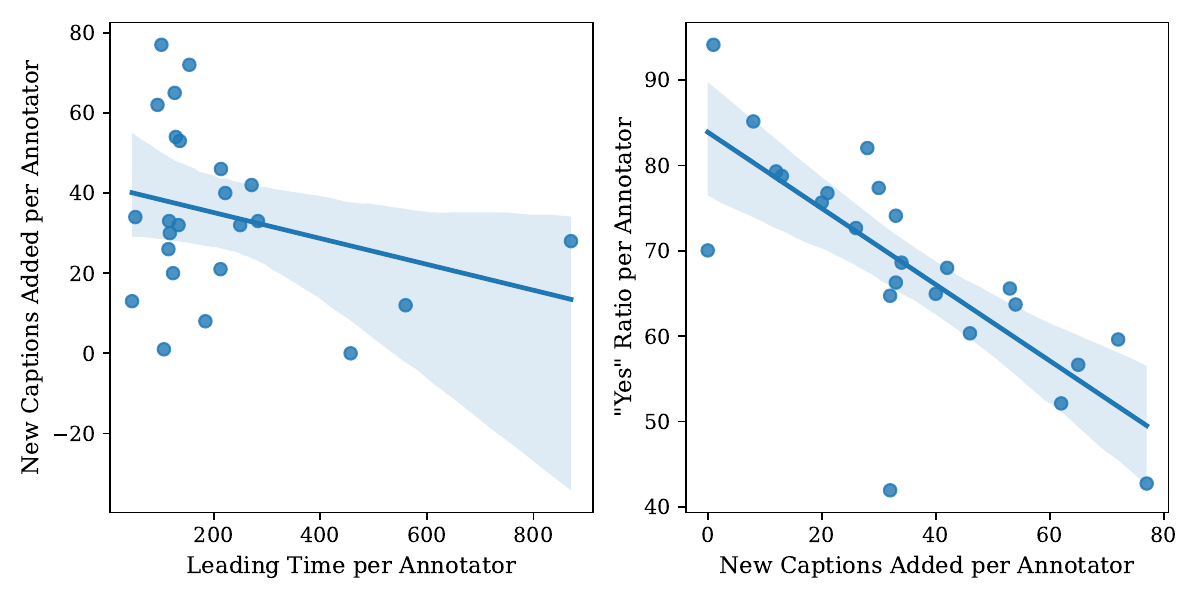}
    \caption{\textbf{Correlation analysis of annotator behavior.} Scatter plots illustrating the relationships between key annotation metrics, overlaid with linear regression lines and confidence intervals. \textbf{Left:} The volume of newly added requests versus the leading time per annotator, showing a slight negative trend. \textbf{Right:} The acceptance rate (``Yes'' ratio) versus the number of newly added request captions. The strong negative correlation indicates that annotators who rejected more simulated requests (lower ``Yes'' ratio) proactively generated more manual requests to ensure dataset quality and completeness.}
    \label{app:human_corr}
\end{figure*}

\begin{figure*}[ht]
    \centering
    \includegraphics[width=0.5\textwidth]{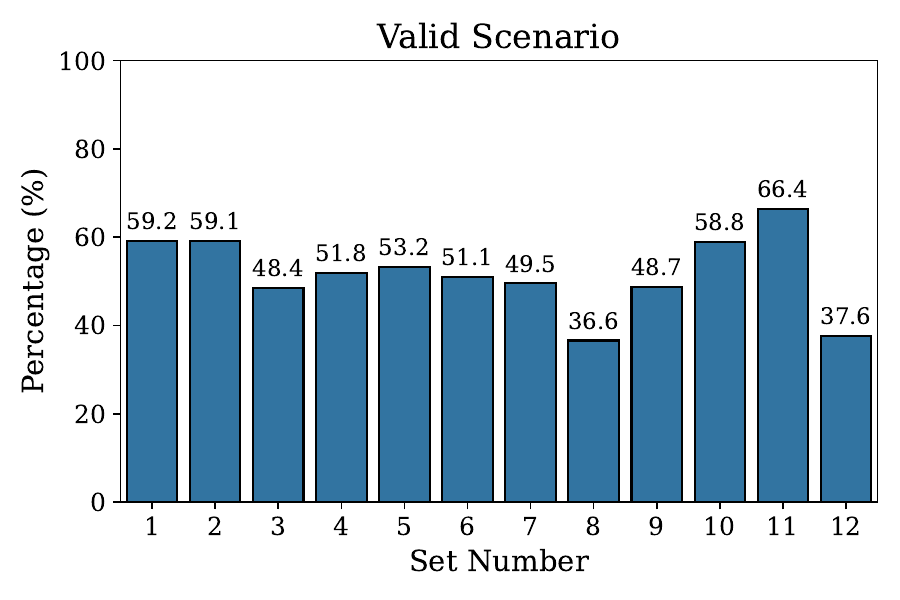}
    \vspace{-1em}
    \caption{\textbf{Percentage of valid requests across annotation sets.} The bar chart displays the proportion of model-generated requests deemed valid by sighted human annotators across 12 distinct evaluation sets. The acceptance rate varies significantly by set, ranging from a low of 36.6\% (Set 8) to a high of 66.4\% (Set 11), highlighting the strictness of the human filtering process required to ensure the quality and safety of the final dataset.}
    \label{app:valid_requests}
\end{figure*}

Participants evaluate each LVLM-generated request with a `yes` or `no` based on its likelihood of being asked by an actual BLV user. If a scene received more than three `no` votes, annotators are required to write and submit new, contextually appropriate requests. The volume of newly added requests and the overall acceptance (`yes`) ratio varied by participant (Fig.~\ref{app:human_dist}, middle and right plots). Notably, a shorter completion time does not correlate with fewer newly added requests, but it is correlated with a higher `yes` ratio (Fig.~\ref{app:human_corr}), suggesting that most annotators strictly adhered to our guidelines. The annotation interface is built using Label Studio~\citep{Label}, resulting in an initial yield of 4,265 (86\%) valid requests (Fig.~\ref{app:valid_requests}).

During post-processing, the authors conduct two additional rounds of manual verification: (1) reviewing generated requests that received only a single `yes` annotation, and (2) evaluating the manually written requests submitted by annotators. Both stages require a consensus between the authors for inclusion. Ultimately, 74 out of 137 marginal generated requests (54\%) and 578 out of 935 annotator-written requests (62\%) are approved, yielding 652 newly verified requests. Furthermore, the authors manually author 62 requests for 12 images that had zero valid requests remaining. This process leads to the final, verified \oursdata dataset totaling 4,979 requests ($4,265 + 652 + 62$).

\subsection{BLV Preferences (\oursdataB)}

\paragraph{Model generation.}

For generating LVLM responses to be evaluated on BLV individuals, we utilize few-shot prompting to extract scene descriptions from five open-source LVLMs~\citep{liu2023visualinstructiontuning, wang2024qwen2vlenhancingvisionlanguagemodels, awadalla2023openflamingoopensourceframeworktraining}, using the samples provided in Tab.~\ref{app:fewshot_prompts_7b_before_outdoors} (outdoor) and Tab.~\ref{app:fewshot_prompts_7b_before_indoors} (indoor). Following generation by these 7B models, we employ GPT-4o mini~\citep{gpt-4omini} to enhance the responses. This refinement process is conducted using the system prompt outlined in Tab.~\ref{app:system_prompts_7b}, and few-shot examples for outdoor (Tab.~\ref{app:fewshot_prompts_gpt_outdoors}) and indoor (Tab.~\ref{app:fewshot_prompts_gpt_indoors}) scenes. For the subsequent second round of BLV user evaluation, we pair a randomly sampled response from one of the 7B models with its GPT-4o mini-enhanced counterpart. Comprehensive details of all system prompts, few-shot examples, and generated response samples are available in Tables~\ref{app:system_prompts_7b}, \ref{app:fewshot_prompts_7b_before_outdoors}/ \ref{app:fewshot_prompts_7b_before_indoors}, and \ref{app:qual_deepcontext_samples}, respectively.

\paragraph{Human study.}

\begin{figure*}[ht]
    \centering
    \includegraphics[width=\textwidth]{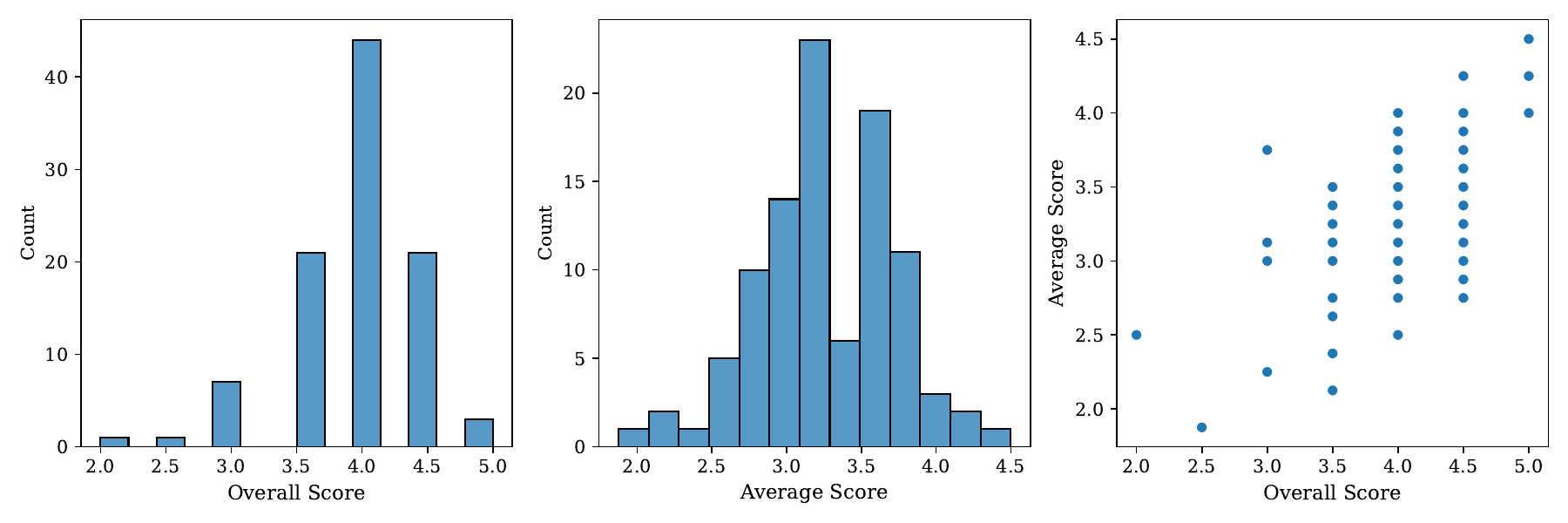}
    \vspace{-1.5em}
    \caption{\textbf{Distribution and correlation of BLV user preference scores.} \textbf{(Left)} Histogram of the overall scores assigned by BLV users, showing a strong peak at 4.0. \textbf{(Middle)} Histogram of the average preference scores across all evaluation criteria, centered around approximately 3.5. \textbf{(Right)} A scatter plot demonstrating a strong positive correlation between the overall score and the averaged criteria score for each data point.}
    \label{app:blv_dist}
\end{figure*}

The initial round of the BLV user experiment requires 1 to 1.5 hours per participant. All eight participants (demographics detailed in Tab.~\ref{tab:personal_blv}) receive compensation of 50,000 KRW ($\sim$34 USD as of January 2025). The duration and compensation for the second experimental round remain identical. The primary distinction between the two rounds is the modality: the first round was conducted offline via direct interview, while the second round was conducted online utilizing screen reader software. Because the first round evaluates specific sub-questions for each image-request pair, the reported ``averaged score'' represents the mean across four distinct categories—Afraidness, Nonactionability, Sufficiency, and Conciseness, which shows positive correlation with the Overall rating (Fig.~\ref{app:blv_dist}). 

\subsection{BLV-Simulated Human Preferences (\oursdataM and \oursdataP)}

\paragraph{Model generation.}

Based on the insight gained from preferences collected directly from BLV users, we generate LVLM responses to be evaluated on sighted human users. After generating baseline responses via in-context learning with the 7B models, we refine the LVLM outputs using GPT-4o mini~\citep{gpt-4omini} based on the prompt in Tab.~\ref{app:system_prompts_ref}. Additionally, Tab.~\ref{app:system_prompts_scoring} details the prompt utilized to generate dimensional scores across seven criteria during the construction of the \oursdata training dataset.

\begin{figure*}[ht]
    \centering
    \includegraphics[width=0.65\textwidth]{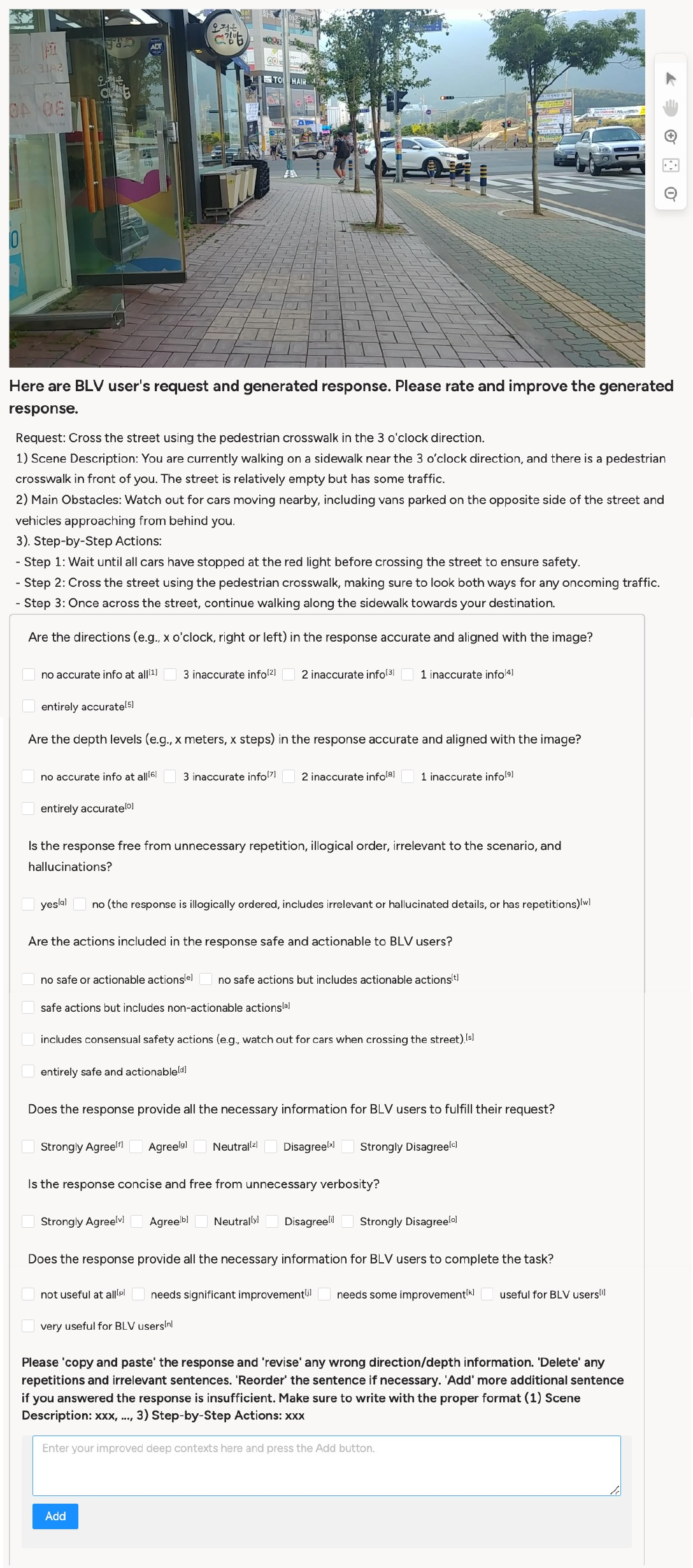}
    \caption{\textbf{Sample screenshot of interface used in the human experiment.} This screen with varying different image-request-response is shown 100 times per annotator.}
    \label{app:screenshot}
\end{figure*}

\paragraph{Human study.}
Annotators are provided with the comprehensive guidelines found in Tab.~\ref{app:guideline} (a sample interface screenshot is available in Fig.~\ref{app:screenshot}). We actively encourage annotators to raise questions regarding technical difficulties or ambiguous edge cases; when queries overlapped, we issue unified clarifications to all annotators to maintain consistency. 

\begin{figure*}[ht]
    \centering
    \includegraphics[width=\textwidth]{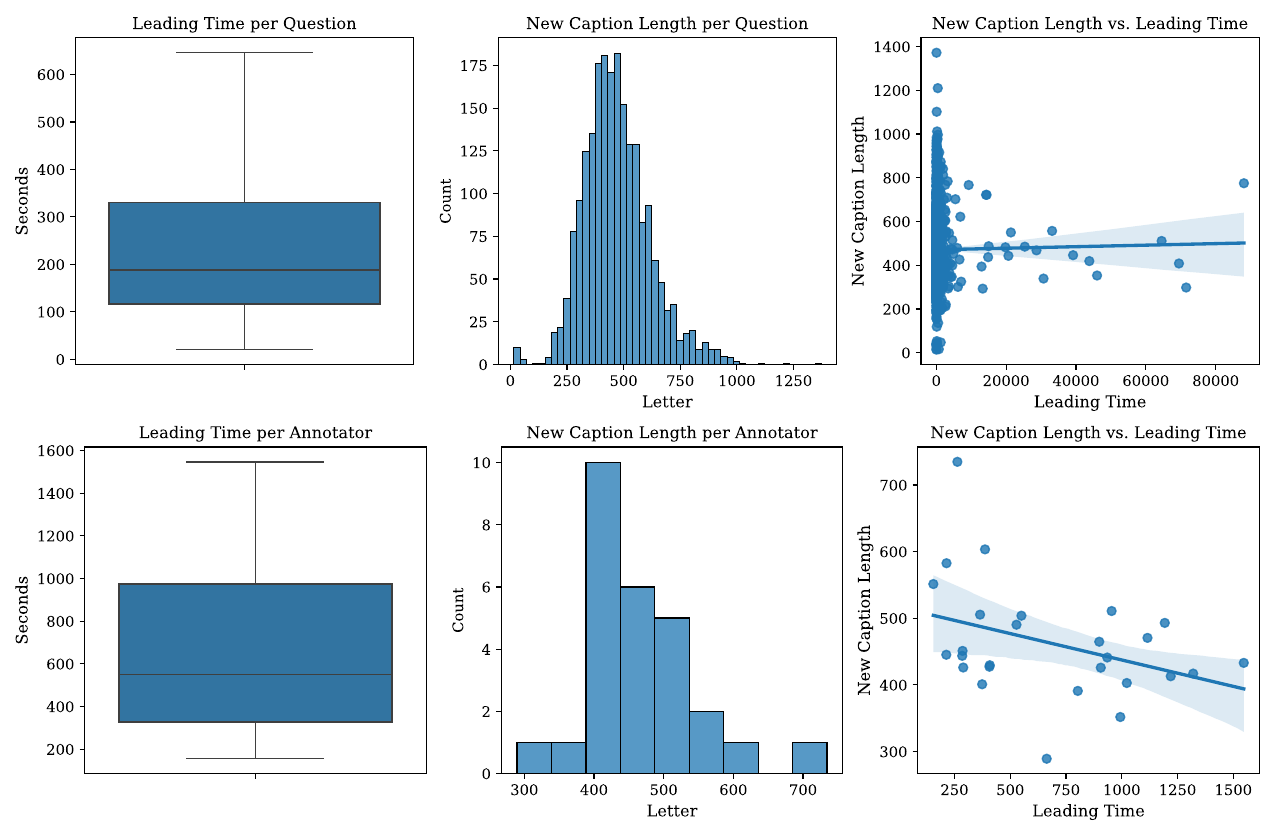}
    \vspace{-2em}
    \caption{\textbf{Distribution and correlation between annotation time per question and the length of newly written human responses.} Longer annotation time does not necessarily correspond to longer or more responses generated by annotators.}
    \label{app:statistics}
\end{figure*}

Task duration and the length of newly added captions are visualized in Fig.~\ref{app:statistics}. The top row visualizes metrics per \emph{question}, while the bottom row visualizes metrics per \emph{annotator}. As demonstrated by the correlation plot (bottom row, third column), a longer annotation time does not strictly correlate with the generation of lengthy captions. 

\paragraph{\oursdata distribution.}

\begin{figure*}[ht]
    \centering
    \includegraphics[width=\textwidth]{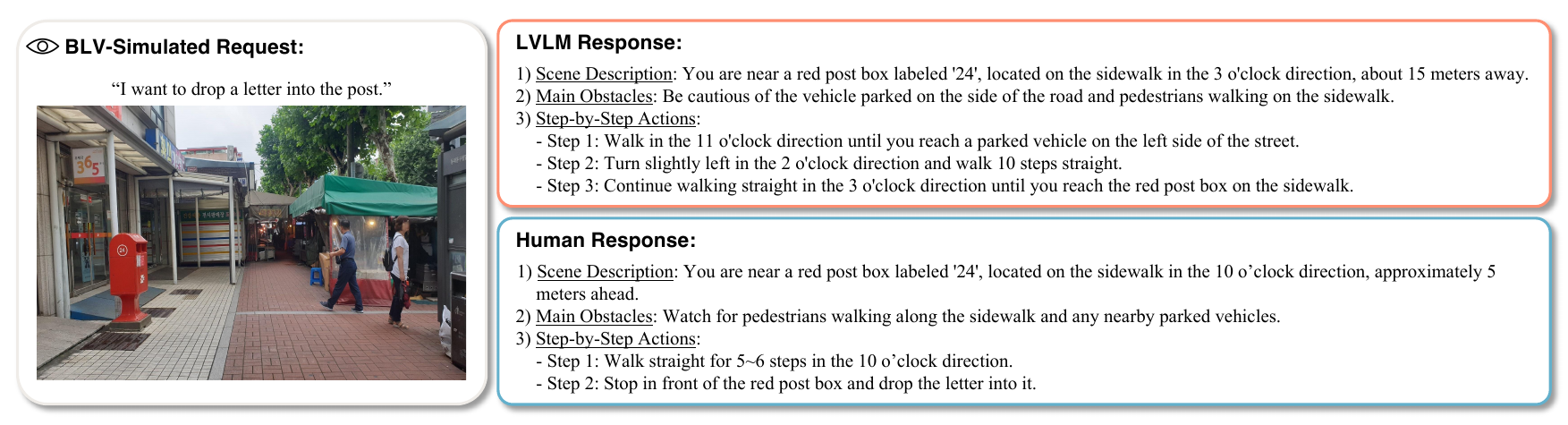}
    \vspace{-1em}
    \caption{\textbf{Example of \oursdataM.} The dataset consists of simulated scenes, user requests, LVLM-generated responses, and scores across seven dimensions: Direction Accuracy, Depth Accuracy, Safety, Sufficiency, Conciseness, and Overall Quality. These scores are normalized and averaged over 2--3 human annotators.}
    \label{app:sample1}
\end{figure*}

\begin{figure*}[ht]
    \centering
    \includegraphics[width=\textwidth]{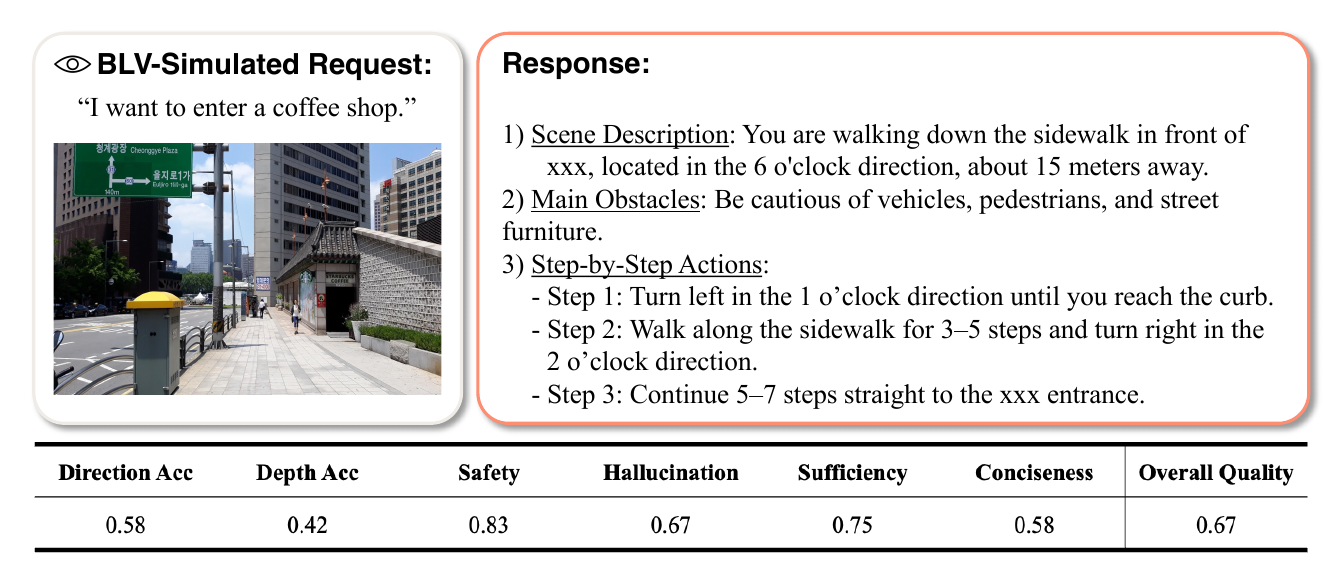}
    \vspace{-1em}
    \caption{\textbf{Example of \oursdataP.} The dataset consists of simulated scenes, user requests, model-generated responses, and human-refined responses. For evaluation, model-generated responses are treated as rejected (negative) samples due to their failure to satisfy all seven criteria outlined in \ref{app:sample1}, whereas human-refined responses are considered chosen (positive) samples, as they meet all criteria.}
    \label{app:sample2}
\end{figure*}

\begin{figure*}[ht]
    \centering
    \includegraphics[width=\textwidth]{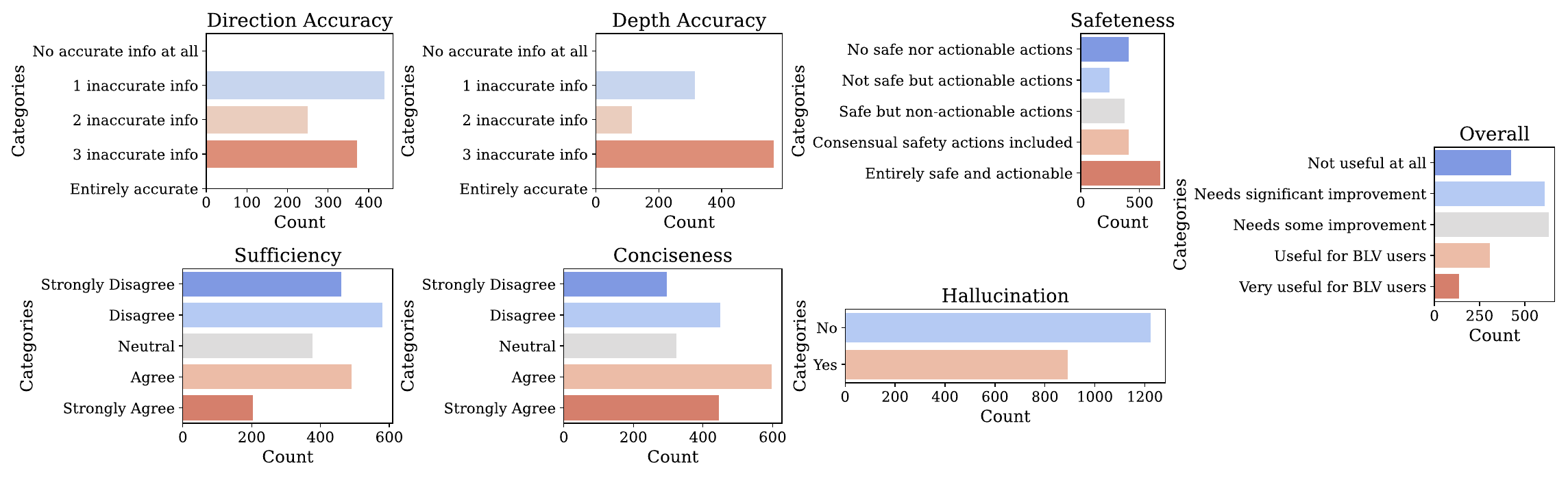}
    \vspace{-1em}
    \caption{\textbf{Distribution of human-annotated scores across seven evaluation criteria.} The bar charts illustrate the baseline performance of LVLM-generated navigational responses. The distributions highlight substantial room for improvement: responses frequently suffer from directional and depth inaccuracies, a high incidence of hallucinations, and an overall rating that heavily skews toward needing significant improvement.} 
    \label{fig:data_dist}
\end{figure*}

Examples of \oursdataM (normalized on a 1--5 scale and averaged across 2--3 annotators) and \oursdataP (pairing positive human texts with negative LVLM responses scoring below 2.5) are provided in Figures~\ref{app:sample1} and \ref{app:sample2}. A summary of the human-annotated scores across all seven evaluation dimensions is presented in Fig.~\ref{fig:data_dist}. We observe that current LVLMs~\citep{xiong2024llava, Qwen2VL, dong2024internlmxcomposer2masteringfreeformtextimage}, including GPT-4o mini~\citep{gpt-4omini}, struggle to consistently generate precise directional and depth information. While most responses are deemed ``entirely safe and actionable,'' LVLM outputs containing even a single spatial inaccuracy pose critical risks for BLV navigation. Improving spatial accuracy remains a primary hurdle for deploying LVLMs in real-world assistive technologies.

\begin{figure}[ht]
    \centering
    \includegraphics[width=0.9\linewidth]{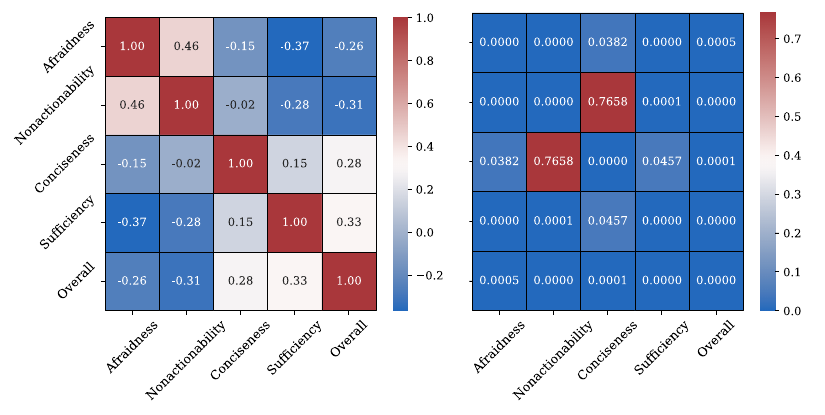}
    \vspace{-0.5em}
    \caption{\textbf{Correlation matrix (and corresponding p-values) of human judgment scores across evaluation criteria.} The matrix displays the pairwise correlation coefficients between different rating categories. The \textit{Overall} quality of the LVLM responses exhibits a stronger absolute correlation with Sufficiency (0.33, \emph{p-value} $< 0.001$), Noncationability (0.31, \emph{p-value} $< 0.001$), indicating these are the primary factors that strongly influence human preference.}
    \label{fig:corr}
\end{figure}

Furthermore, annotators frequently disagree with the assertion that LVLM responses are \emph{sufficient}, contrasting with a higher consensus on \emph{conciseness}. Although sufficiency is inherently subjective (dependent on whether the annotator feels all necessary navigational context is provided), it exhibits the strongest correlation with the overall human rating (0.85, \emph{see} Fig.~\ref{fig:corr}). This indicates that response sufficiency is the main driver of perceived quality. Conversely, the \emph{hallucination} category shows the lowest correlation (0.35), likely due to differing scoring scales. However, with nearly 1,000 responses containing identified hallucinations, this failure mode remains prevalent. Ultimately, these distributions underscore the urgent need to enhance the ability of LVLMs to generate comprehensive, accurate, and task-relevant context for BLV users.

\begin{table*}[ht]
\centering
\vspace{1em}
\begin{tcolorbox}[
    enhanced,
    colback=black!2, 
    colframe=black!70, 
    colbacktitle=black!10, 
    coltitle=black, 
    fonttitle=\bfseries\small, 
    title=System Prompt for Generating BLV-Simulated Requests,
    boxsep=4pt, 
    left=6pt, 
    right=6pt, 
    top=4pt, 
    bottom=4pt, 
    arc=3pt, 
    fontupper=\small, 
    fontlower=\small
]
You are a request writer. Given an image, your task is to generate 5 to 10 requests related to actions that blind or low-vision (BLV) people can perform. Each request must describe specific, actionable tasks in a detailed and structured manner. The focus should be on mobility, particularly actions related to safe movement, object manipulation, or accessing information that BLV individuals can perform within the context of the scene. The requests should not overlap with each other but be diverse, detailed, and read-recognized texts. Please do not mention an object or person not detected in the image, and refrain from using unclear or useless verbs such as organize, explore, navigate, locate, feel, check, and gather information. Do not include color and auditory information. The output should be properly formatted as a list containing 5 to 10 requests.
\end{tcolorbox}
\vspace{-0.5em}
\caption{\textbf{System prompt for simulating BLV mobility requests.} We instruct GPT-4o mini to generate 5 to 10 diverse, actionable, and visually grounded requests per image. The prompt explicitly constrains the model to avoid auditory or color cues, hallucinated objects, and vague verbs, ensuring the generated tasks are realistic for BLV users.}
\label{app:system_prompts_scengen}
\end{table*}
\begin{table*}[ht]
\centering
\begin{tcolorbox}[
    enhanced,
    colback=black!2, 
    colframe=black!70, 
    colbacktitle=black!10, 
    coltitle=black, 
    fonttitle=\bfseries\small, 
    title=Few-Shot Prompt Sample for Generating BLV-Simulated Requests (Outdoor Scenes),
    boxsep=4pt, 
    left=6pt, 
    right=6pt, 
    top=4pt, 
    bottom=4pt, 
    arc=3pt, 
    segmentation style={solid, draw=black!30, line width=0.5pt}, 
    fontupper=\small, 
    fontlower=\small
]
Here is an example. This is a sample image. Based on the given image, you can give requests as follows:\vspace{1ex}

\hspace*{1em}\textbf{1.} Reach the other side of the street.\\
\hspace*{1em}\textbf{2.} Enter the market named xxx.\\
\hspace*{1em}\textbf{3.} Enter the parking lot marked by xxx sign named xxx.\\
\hspace*{1em}\textbf{4.} Go in front of a signboard with parking information (30xxx).\\
\hspace*{1em}\textbf{5.} Enter the karaoke place named xxx in the opposite side building identified by red signs.\\
\hspace*{1em}\textbf{6.} Enter the place called xxx.\\
\hspace*{1em}\textbf{7.} Go towards the tall tree.\vspace{1.5ex}

\textbf{You should not suggest requests as follows:}\vspace{1ex}

\hspace*{1em}\textbf{1.} Enter the store. \textcolor{deepred}{\emph{(\underline{unspecific information}; If there are multiple stores, you should point out more specific stores with names, if possible.)}} \\
\hspace*{1em}\textbf{2.} Enjoy the scenery of the street view. \textcolor{deepred}{\emph{(\underline{uninformative information}; Avoid the usage of xxx.)}} \\
\hspace*{1em}\textbf{3.} Pick up the trash and throw it in the trash can that is located 3 steps away in the 2 o'clock direction. \textcolor{deepred}{\emph{(\underline{useless request for BLV users}; Avoid generating request not related with mobility.)}} \\
\hspace*{1em}\textbf{4.} Avoid pedestrian(s) walking toward me. \textcolor{deepred}{\emph{(\underline{fake request}; No pedestrian in image.)}} \\
\hspace*{1em}\textbf{5.} Approach the signpost marked with `P'. \textcolor{deepred}{\emph{(\underline{ambiguous request}; If there are same multiple objects with different direction, you should specify more detailed, if possible.)}} \\
\hspace*{1em}\textbf{6.} Feel the texture of the brick wall on your right. \textcolor{deepred}{\emph{(\underline{useless request}; Avoid generating request without mobility concept.)}} \\
\hspace*{1em}\textbf{7.} Take a picture of sign post on your left for reading. \textcolor{deepred}{\emph{(\underline{unspecific information}; If there are multiple sign posts, you should specify more detailed, if possible.)}} \\
\hspace*{1em}\textbf{8.} Stand near the bushes in the middle of the road. \textcolor{deepred}{\emph{(\underline{useless request}; This is not a mobility request).}}
\end{tcolorbox}
\vspace{-0.5em}
\caption{\textbf{Few-shot prompt example for outdoor request generation.} We provide GPT-4o mini with a sample image alongside a curated list of acceptable and unacceptable BLV mobility requests. The negative examples explicitly outline common generation errors to guide the model toward high-quality outputs (Note: xxx represents store names or signs detected in the original image).}
\vspace{-1em}
\label{app:fewshot_prompts_scengen_outdoors}
\end{table*}
\begin{table*}[ht]
\centering
\begin{tcolorbox}[
    enhanced,
    colback=black!2, 
    colframe=black!70, 
    colbacktitle=black!10, 
    coltitle=black, 
    fonttitle=\bfseries\small, 
    title=Few-Shot Prompt Sample for Generating BLV-Simulated Requests (Indoor Scenes),
    boxsep=4pt, 
    left=6pt, 
    right=6pt, 
    top=4pt, 
    bottom=4pt, 
    arc=3pt, 
    fontupper=\small, 
    fontlower=\small
]
Here is an example. This is a sample image. Based on the given image, you can give requests as follows:\vspace{1ex}

\hspace*{1em}\textbf{1.} Pick up the red fruits.\\
\hspace*{1em}\textbf{2.} Enter the room.\\
\hspace*{1em}\textbf{3.} Go outside the apartment.\\
\hspace*{1em}\textbf{4.} Clean the pots on the stove.\\
\hspace*{1em}\textbf{5.} Pick up the plant pot.\\
\hspace*{1em}\textbf{6.} Water the plants right of the refrigerator.\\
\hspace*{1em}\textbf{7.} Sit on the chair in the kitchen.\\
\hspace*{1em}\textbf{8.} Check the gas valve is securely locked.\vspace{1.5ex}

\textbf{You should not suggest requests as follows:}\vspace{1ex}

\hspace*{1em}\textbf{1.} Look at yourself in a mirror in the 12 o'clock direction. \textcolor{deepred}{\emph{(\underline{useless information}; BLV users cannot see themselves.)}} \\
\hspace*{1em}\textbf{2.} Pick up a phone on the table on the right. \textcolor{deepred}{\emph{(\underline{misidentified object}; There is no phone on the table.)}}\\
\hspace*{1em}\textbf{3.} Go to the kitchen. \textcolor{deepred}{\emph{(\underline{useless and undetailed information}; The user is already located in the kitchen.)}}\\
\hspace*{1em}\textbf{4.} Open the door. \textcolor{deepred}{\emph{(\underline{unspecific information}; There are multiple doors.)}}\\
\hspace*{1em}\textbf{5.} Organize the items on the table. \textcolor{deepred}{\emph{(\underline{unspecific information}; There are multiple items on the table.)}}\\
\hspace*{1em}\textbf{6.} Clear any debris on the floor. \textcolor{deepred}{\emph{(\underline{unspecific information}; You should specify the location and object more detailed.)}}\\
\hspace*{1em}\textbf{7.} Open the door. \textcolor{deepred}{\emph{(\underline{unspecific information}; There are multiple doors.)}}\\
\hspace*{1em}\textbf{8.} Feel the texture of the fruits on the table. \textcolor{deepred}{\emph{(\underline{useless information}; Avoid generating requests without the mobility concept.)}}
\end{tcolorbox}
\vspace{-0.5em}
\caption{\textbf{Few-shot prompt example for indoor request generation.} Similar to the outdoor setting, we provide GPT-4o mini with an indoor sample image and a curated list of acceptable and unacceptable BLV mobility requests. The negative examples explicitly highlight common indoor generation errors, providing nonspecific directions or suggesting visually dependent actions.}
\label{app:fewshot_prompts_scengen_indoors}
\end{table*}

\begin{table*}[ht]
\centering
\begin{tcolorbox}[
    enhanced,
    colback=black!2, 
    colframe=black!70, 
    colbacktitle=black!10, 
    coltitle=black, 
    fonttitle=\bfseries\small, 
    title=Human Labeling Guidelines for Evaluating BLV-Simulated Requests,
    boxsep=4pt, 
    left=6pt, 
    right=6pt, 
    top=4pt, 
    bottom=4pt, 
    arc=3pt, 
    segmentation style={solid, draw=black!30, line width=0.5pt}, 
    fontupper=\small, 
    fontlower=\small
]
\textbf{Introduction}: The goal of this labeling task is to create a dataset that provides detailed and actionable descriptions of mobility requests related to blind or low-vision (BLV) users. The requests gathered will ultimately support BLV users in safely navigating and interacting with their surroundings, both indoors and outdoors. As an annotator, your task is to review 5 to 10 requests associated with a given image and decide if each request accurately reflects the actions a BLV person could take in that environment. You will respond with `Yes' or `No' for each request.\vspace{1.5ex}

\textbf{Image Types}: Each image depicts an indoor or outdoor setting, such as a street, living room, or public space. Your role is to assess mobility-related actions a BLV user could realistically perform within that context.\vspace{1.5ex}

\textbf{Request Concept}:\\
A valid request should describe specific and actionable tasks that a BLV user might perform within the scene. For example, requests should focus on mobility actions, such as safe movement, spatial exploration, or interaction with objects. Requests must be directly related to observable objects in the image and provide clear, specific details. (\emph{\eg}, ``Approach the door on the right,'' ``Pick up a book from the desk in front of you,'' ``Organize the books on the table.'').\vspace{1ex}

Invalid requests include:\vspace{0.5ex}

\hspace*{1em}(1) Requests that involve objects not visible in the image.\\
\hspace*{1em}(2) Requests requiring actions based on color or auditory cues, which are difficult for a BLV person to perceive.\\
\hspace*{1em}(3) Requests that are vague or ambiguous, such as when identical objects appear in different locations, but the request doesn't specify which object is referenced (\emph{\eg}, if there are two tables with water bottles on the left and right and the request says, ``Pick up the water bottle from the table,'' this is unclear and should be marked ``No''). Conversely, if the request says, ``Pick up the water bottle from the table on the user's right,'' it is clear and should be marked ``Yes.''\\
\hspace*{1em}(4) Requests with unclear descriptions. (\emph{\eg}, ``Feel the texture of the right wall,'' ``Explore the bushes on the left.'')\\
\hspace*{1em}(5) If an object in the request is not clearly identified, select ``No'' (\emph{\eg}, if there are many items on a shelf and the request says, ``Organize the items on the shelf,'' it should be marked ``No'').\vspace{1.5ex}

\textbf{Annotation Task}: For each image, read 5 to 10 provided requests and determine if they align with the request concept outlined above. If a request is appropriate, select ``Yes.'' If not, select ``No.'' You may also suggest new requests, provided they do not overlap with the given ones. When reviewing requests, keep the following criteria in mind:\vspace{1ex}

\hspace*{1em}\textbullet~ \emph{Specificity}: Does the request offer specific details? For example, if a request suggests ``Enter a store,'' but multiple stores are visible, it should clearly indicate which one.\\
\hspace*{1em}\textbullet~ \emph{Relevance}: Is the request mobility-focused and actionable? Avoid sensory-focused requests (\eg, ``Feel the texture'') or vague references to general actions.\\
\hspace*{1em}\textbullet~ \emph{Accuracy}: Ensure the request doesn't reference objects, people, or actions not visible in the image (\eg, mentioning cars or items that aren't present).\\
\hspace*{1em}\textbullet~ \emph{Clarity}: Avoid ambiguous verbs like ``explore,'' ``organize,'' or ``inspect.'' Requests should describe clear, executable actions.\vspace{1ex}

If you select more than 3 ``No'', you should propose more than one request related to BLV mobility if you believe they are relevant:\vspace{0.5ex}

\hspace*{1em}(1) Be specific. Use precise details such as ``3 o'clock direction,'' ``12 steps ahead,'' or ``door with an exit sign.''\\
\hspace*{1em}(2) Focus on mobility: prioritize actions related to movement, positioning, or object interaction.\\
\hspace*{1em}(3) Only mention objects and people visible in the image.\\
\hspace*{1em}(4) Ensure that the request describes an action that a BLV user can realistically perform.\\
\hspace*{1em}(5) Avoid vague or unrelated verbs like ``explore,'' ``examine,'' or ``feel.''\\
\hspace*{1em}(6) Do not reference objects not present in the image (\eg, unseen cars, people, or signs).\\
\hspace*{1em}(7) Avoid suggesting actions that are out of context, such as ``enjoy nature'' or ``wait for the bus'' if no bus stop is visible.\vspace{1.5ex}

\textbf{Request Examples}: [refer to few-shot samples]\vspace{1.5ex}

\textbf{Final Note}: The ultimate goal of this task is to collect realistic and helpful requests for BLV users. Your annotations will help ensure the requests reflect actions that can be performed in real-life situations by BLV users.
\end{tcolorbox}
\vspace{-0.5em}
\caption{\textbf{Human experiment guideline for evaluating and refining BLV mobility requests.} The guidelines detail the workflow for validating simulated requests.}
\label{app:request_guideline}
\end{table*}


\begin{table*}[ht]
\centering
\begin{tcolorbox}[
    enhanced,
    colback=black!2, 
    colframe=black!70, 
    colbacktitle=black!10, 
    coltitle=black, 
    fonttitle=\bfseries\small, 
    title=Few-Shot Examples for Generating LVLM Responses (Outdoor Scenes),
    boxsep=4pt, 
    left=6pt, 
    right=6pt, 
    top=4pt, 
    bottom=4pt, 
    arc=3pt, 
    segmentation style={solid, draw=black!30, line width=0.5pt}, 
    fontupper=\small, 
    fontlower=\small
]
\textbf{Request:} Enter the store named xxx.\vspace{1ex}

\textbf{Response (\emph{before}):}\\
1. The store named xxx is located in the 10 o'clock direction, approximately 2 to 3 meters away. Be aware that a signpost is in the 9 to 10 o'clock direction, about two to three steps away. 2. Turn left in the 10 o'clock direction and walk straight for three to five steps. Walk with caution as you are on a roadway. There is a parked car about ten steps ahead of you, and the cars may be exiting a building on your right, approximately 4 to 5 meters away. 3. Next, turn left in the 11 o'clock direction. Be mindful of a very low curb and a step near the store entrance. 4. The door appears to be closed, so you will need to open it to enter the store.\vspace{1ex}

\textbf{Response (\emph{after}):}\\
\hspace*{1em}\textbf{1) Scene Description:} You are on a roadway near a store named xxx, located in the 10 o'clock direction, about 2--3 meters away.\\
\hspace*{1em}\textbf{2) Main Obstacles:} There is a signpost located on the front-left side of the store, and there is a low curb and a step near the store entrance.\\
\hspace*{1em}\textbf{3) Step-by-Step Actions:}\\
\hspace*{2em}-- Step 1: Turn left in the 10 o'clock direction and walk 3--5 steps, avoiding the signpost on your left after the turn.\\
\hspace*{2em}-- Step 2: Turn left again in the 11 o'clock direction and walk 2--3 steps to enter the store.

\tcblower

\textbf{Request:} Enter the market with a red sign (named xxx).\vspace{1ex}

\textbf{Response (\emph{before}):}\\
1. The market with a red sign (named xxx) is located in the 11 o'clock direction, approximately 12 meters away. Be cautious of multiple obstacles, including stacked boxes on your left and parked cars on your right, as you proceed to xxx due to ongoing construction. 2. To start, walk about five to seven steps in the 1 o'clock direction. As you continue, be aware of a parked SUV on the left. 3. Next, turn slightly left in the 11 to 12 o'clock direction and walk straight for five steps, being cautious of parked cars and pedestrians. There is also the roadway on your right. 4. Finally, walk straight for ten steps, then turn left. Walk another five to eight steps to reach the store entrance. 5. There are several parked cars in front of the store, so proceed carefully.\vspace{1ex}

\textbf{Response (\emph{after}):}\\
\hspace*{1em}\textbf{1) Scene Description:} You are near a market with a red sign named ``xxx'', located in the 11 o'clock direction, approximately 12 meters away.\\
\hspace*{1em}\textbf{2) Main Obstacles:} Be cautious of stacked boxes, a parked SUV on the left side of the street, and pedestrians and parked trucks on the right side of the street.\\
\hspace*{1em}\textbf{3) Step-by-Step Actions:}\\
\hspace*{2em}-- Step 1: Walk 5--7 steps in the 1 o'clock direction until you reach a parked SUV on your left side.\\
\hspace*{2em}-- Step 2: Turn slightly left in the 11--12 o'clock direction and walk 10 steps straight.\\
\hspace*{2em}-- Step 3: Continue 5--8 steps to the store entrance, avoiding several parked cars near the entrance.

\tcblower

\textbf{Request:} Enter the market named xxx.\vspace{1ex}

\textbf{Response (\emph{before}):}\\
1. The market is located across the street in the 12 o'clock direction, approximately 9 to 10 meters away. To reach it, you will need to cross the road, so proceed with caution. 2. First, turn left in the 9 to 10 o'clock direction and cross the street until you reach the curb. 3. Turn right in the 3 o'clock direction and walk straight for 15 to 20 steps. There are multiple stores, including a church and a real estate building on your left, so please watch out for pedestrians exiting or entering these stores. 4. Once you feel a hard object with your cane or foot, slightly turn left in the 11 o'clock direction and walk three to five steps to enter the store.\vspace{1ex}

\textbf{Response (\emph{after}):}\\
\hspace*{1em}\textbf{1) Scene Description:} You are near a market named ``xxx'' located across the street at the 12 o'clock direction, about 9--10 meters away.\\
\hspace*{1em}\textbf{2) Main Obstacles:} Watch out for pedestrians near stores near xxx.\\
\hspace*{1em}\textbf{3) Step-by-Step Actions:}\\
\hspace*{2em}-- Step 1: Turn left toward the 9--10 o'clock direction until you reach the curb.\\
\hspace*{2em}-- Step 2: Turn right in the 3 o'clock direction and walk 15--20 steps straight along the sidewalk.\\
\hspace*{2em}-- Step 3: Turn left in the 11 o'clock direction and walk 3--5 steps to reach the xxx entrance.
\end{tcolorbox}
\vspace{-0.5em}
\caption{\textbf{Few-shot prompt examples for LVLM response generation in outdoor scene contexts.} These examples demonstrate the transformation from unstructured, conversational outputs (\emph{before}) to highly structured, actionable navigational guidance (\emph{after}). The \emph{after} format strictly separates scene context, obstacle warnings, and quantitative step-by-step instructions (Note: xxx represents store names or signs detected in the original image).}
\vspace{-1em}
\label{app:fewshot_prompts_7b_before_outdoors}
\end{table*}
\begin{table*}[ht]
\centering
\begin{tcolorbox}[
    enhanced,
    colback=black!2, 
    colframe=black!70, 
    colbacktitle=black!10, 
    coltitle=black, 
    fonttitle=\bfseries\small, 
    title=Few-Shot Examples for Generating LVLM Responses (Indoor Scenes),
    boxsep=4pt, 
    left=6pt, 
    right=6pt, 
    top=4pt, 
    bottom=4pt, 
    arc=3pt, 
    segmentation style={solid, draw=black!30, line width=0.5pt}, 
    fontupper=\small, 
    fontlower=\small
]
\textbf{Request:} Pick up the red fruits.\vspace{1ex}

\textbf{Response (\emph{before}):}\\
1. The red fruits are on the table directly in front of you, slightly to the right, at the 2 o'clock position. 2. To locate them, place your hand at about a 30-degree angle from straight ahead and feel the edge of the table. 3. Slowly trace the edge of the table to your right, and be mindful of a chair in front of the table. 4. The red fruits are in a dish next to a tissue box. Be cautious, as there is a stove with pots near the table.\vspace{1ex}

\textbf{Response (\emph{after}):}\\
\hspace*{1em}\textbf{1) Scene Description:} You are in a kitchen with a table located directly in front of you at the 2 o'clock direction.\\
\hspace*{1em}\textbf{2) Main Obstacles:} There is a chair without a backrest and a stove with pots located not directly in front of you but on the opposite side of the table.\\
\hspace*{1em}\textbf{3) Step-by-Step Actions:}\\
\hspace*{2em}-- Step 1: Place your right hand in the 2 o'clock direction to locate the edge of the table.\\
\hspace*{2em}-- Step 2: Slowly trace the edge of the table until you take two right turns.\\
\hspace*{2em}-- Step 3: You will find red fruits directly after your last turn in front of a tissue box.

\tcblower

\textbf{Request:} Sit on the sofa.\vspace{1ex}

\textbf{Response (\emph{before}):}\\
1. The sofa is located in the 10 to 11 o'clock direction. Be mindful of a low wooden table with a plant pot and a globe between you and the sofa, requiring you to navigate around it. 2. To do so, turn slightly to the right, toward the 12 to 1 o'clock direction, and walk three to five steps. There is a carpet, so be careful not to slip. 3. Then, turn 90 degrees left and walk two to three steps to reach the left end of the sofa. You can choose to sit on the left end or move to the right side of the sofa. If you move to the right, be cautious of the low wooden table in front of the sofa. 4. Additionally, there is a massage chair to your right. Watch out for objects such as dumbbells and plant pots placed on the low wooden table.\vspace{1ex}

\textbf{Response (\emph{after}):}\\
\hspace*{1em}\textbf{1) Scene Description:} You are in a living room with a sofa located about 3 to 4 meters away in the direction of 10--11 o'clock.\\
\hspace*{1em}\textbf{2) Main Obstacles:} There is a low wooden table directly in front of the sofa with the same width as the sofa.\\
\hspace*{1em}\textbf{3) Step-by-Step Actions:}\\
\hspace*{2em}-- Step 1: Turn right in the 12--1 o'clock direction, and walk 3--5 steps until you reach the end of the low wooden table.\\
\hspace*{2em}-- Step 2: Turn 90 degrees to your left and walk 2--3 steps to reach the left end of the sofa.

\tcblower

\textbf{Request:} Pick up the book on 4th floor of the 5-story wooden bookshelf.\vspace{1ex}

\textbf{Response (\emph{before}):}\\
1. The 5-story wooden bookshelf is located in the 10 o'clock direction, about 6 meters away. However, there are obstacles between you and the bookshelf, so you'll need to navigate carefully. 2. First, turn left toward the 9 to 10 o'clock direction and walk straight for five to eight steps. As you proceed, be cautious of a 1-person sofa and a wooden table and chair on your right. 3. After you reach the wooden table and chair, turn 90 degrees right and walk three to four steps to reach the wooden bookshelf. 4. The shelf holds multiple books, but be cautious of objects like a plant pot placed on the very top of the bookshelf.\vspace{1ex}

\textbf{Response (\emph{after}):}\\
\hspace*{1em}\textbf{1) Scene Description:} You are in a room with a 5-story wooden bookshelf located at the 10 o'clock direction, about 6 meters away.\\
\hspace*{1em}\textbf{2) Main Obstacles:} There is a low circular table with a 1-person sofa and a chair in the 2 o'clock direction about 2 to 3 steps away, and there is also another set of tables and chairs directly in front-right of the 5-story wooden bookshelf.\\
\hspace*{1em}\textbf{3) Step-by-Step Actions:}\\
\hspace*{2em}-- Step 1: Turn left toward the 9--10 o'clock direction and walk straight for 5--8 steps.\\
\hspace*{2em}-- Step 2: When you reach the set of wooden table and chairs on your right and a low leather sofa on your left, turn 90 degrees to your right.\\
\hspace*{2em}-- Step 3: Walk 3--4 steps to reach the 5-story wooden bookshelf.
\end{tcolorbox}
\vspace{-0.5em}
\caption{\textbf{Few-shot prompt examples for LVLM response generation in indoor scene contexts.} Similar to the outdoor setting, these examples guide the 7B models to transition from unstructured, conversational descriptions (\emph{before}) to a strict, three-part format (\emph{after}). The refined format strictly isolates the room's context, indoor obstacles, and explicit step-by-step physical actions.}
\label{app:fewshot_prompts_7b_before_indoors}
\end{table*}

\begin{table*}[ht]
\centering
\begin{tcolorbox}[
    enhanced,
    colback=black!2, 
    colframe=black!70, 
    colbacktitle=black!10, 
    coltitle=black, 
    fonttitle=\bfseries\small, 
    title=System Prompts for Generating LVLM Responses (\emph{Before} and \emph{After}),
    boxsep=4pt, 
    left=6pt, 
    right=6pt, 
    top=4pt, 
    bottom=4pt, 
    arc=3pt, 
    segmentation style={solid, draw=black!30, line width=0.5pt}, 
    fontupper=\small, 
    fontlower=\small
]
\textbf{7B LVLMs}:\vspace{1ex}

You are an expert at evaluating the quality of the model responses for a given task. The task for the model was to assist Blind and Low-Vision (BLV) users by providing them with details for their text-based simple requests, given an image of a visual scene. You will be given an image of a visual scene, the text-based request, and the text-based model response. Enhance the response in terms of three criteria:\vspace{1ex}

\hspace*{1em}\textbf{a) Accuracy:} Your response should include correct information of direction (\eg, x o'clock) and depth (\eg, x steps).\\
\hspace*{1em}\textbf{b) Length:} Your response should include all the correct detailed information; however, it should not include useless information for BLV users, such as color or non-existent assumptions.\\
\hspace*{1em}\textbf{c) Actionability:} Your response should only include possible safe actions that the BLV user can perform and include useful objects, such as braille blocks, and cautious objects, such as motorcycles, that might come toward the BLV user.\vspace{1.5ex}

\textbf{GPT-4o mini}:\vspace{1ex}

You are an expert at providing a Blind or Low Vision (BLV) an accurate, helpful description, given an environmental scene (outdoor or indoor) and corresponding to their text-based request. Remember that BLV users cannot see as much as normally-sighted humans, so you must provide detailed but precise information from the image. Enhance the model response by including precise clock directions (options: 9, 10, 11, 12, 1, 2, 3 o'clock), depth levels (in meters or steps), and objects BLV users should avoid or utilize. 

\tcblower

\textbf{7B LVLMs}:\vspace{1ex}

You are an expert at providing responses for blind or low-vision (BLV) users. Given an indoor or outdoor visual scene photo taken by a BLV user and their text-based requests, your task is to respond to user requests with accurate, structured, and actionable responses. Please ensure your response includes the following:\vspace{1.5ex}

\hspace*{1em}\textbf{1) Scene Description:} Provide a single, concise sentence describing the scene or environment relevant to the request, including precise directions (0 to 90 degrees left or right) and depths in meters. Do not include any color information.\\
\hspace*{1em}\textbf{2) Main Obstacles:} Highlight only one or two potential challenges/obstacles in a single sentence. Avoid generic, obvious warnings, and do not include non-detected obstacles.\\
\hspace*{1em}\textbf{3) Step-by-step Actions:} Outline the key actions required to fulfill the request in 1-3 clear and concise sentences. Include precise directions (9 to 3 o'clock directions) and depths in steps (\eg, 3-5 steps) for each stage.
\end{tcolorbox}
\vspace{-0.5em}
\caption{\textbf{Evolution of system prompts for LVLM response generation.} The top section displays the initial prompts provided to the 7B LVLMs and GPT-4o mini, which primarily focused on open-ended refinement. The bottom section presents the finalized, highly structured prompt for the 7B LVLMs, designed to strictly enforce a three-part response format (Scene Description, Obstacles, and Actions) while mandating specific navigational formats like clock directions and step depths.}
\label{app:system_prompts_7b}
\end{table*}

\begin{table*}[ht]
\centering
\begin{tcolorbox}[
    enhanced,
    colback=black!2, 
    colframe=black!70, 
    colbacktitle=black!10, 
    coltitle=black, 
    fonttitle=\bfseries\small, 
    title=Few-Shot Examples for Refining LVLM Responses Using GPT-4o mini (Outdoor Scenes),
    boxsep=4pt, 
    left=6pt, 
    right=6pt, 
    top=4pt, 
    bottom=4pt, 
    arc=3pt, 
    segmentation style={solid, draw=black!30, line width=0.5pt}, 
    fontupper=\small, 
    fontlower=\small
]
\textbf{Request:} Enter the cafe named xxx.\vspace{1ex}

\textbf{Response:}\\
\hspace*{1em}\textbf{1.} The cafe named xxx is located in the 3 o'clock direction.\\
\hspace*{1em}\textbf{2.} First, turn right in the direction of 2 o'clock and walk six to eight steps.\\
\hspace*{1em}\textbf{3.} Then, turn left in the 9 o'clock direction and walk two to three steps to reach the cafe's entrance.\\
\hspace*{1em}\textbf{4.} However, there are some obstacles between you and the cafe, so please be careful.\vspace{1ex}

\textbf{Enhanced response:}\\
\hspace*{1em}\textbf{1.} The entrance of the cafe named xxx is located at 2 o'clock direction, approximately 10 meters away. You are currently on the road in an alley. Be cautious of an approaching black car in the 12 o'clock direction approximately 15 meters away. You could turn right and walk more than 5 steps to avoid the car.\\
\hspace*{1em}\textbf{2.} Turn left again to face the xxx in the 12 to 1 o'clock direction.\\
\hspace*{1em}\textbf{3.} Walk 10 to 15 steps straight; however, watch out for the billboard on your left as you walk in the earlier steps. Also, slightly turn left in the 11 to 12 o'clock direction since there is a low curb on your right in the later steps.\\
\hspace*{1em}\textbf{4.} Walk straight for three to five steps, and turn right in the 3 o'clock direction.\\
\hspace*{1em}\textbf{5.} The cafe is 2 meters away; however, there is a parked car in front of the entrance, so proceed carefully.

\tcblower

\textbf{Request:} Approach the subway station exit with the number 4.\vspace{1ex}

\textbf{Response:}\\
\hspace*{1em}\textbf{1.} The subway station exit with the number 4 is on your right. There might be obstacles between you and the exit, so please be careful.\\
\hspace*{1em}\textbf{2.} First, turn right at 2 o'clock and walk five to seven steps.\\
\hspace*{1em}\textbf{3.} Then, turn right slightly and proceed ten to twelve steps. There are stairs in front of the exit, so be careful.\vspace{1ex}

\textbf{Enhanced response:}\\
\hspace*{1em}\textbf{1.} The subway station exit with the number 4 is located to your right in the 2 o'clock direction, approximately 15 meters away, on the opposite side of the street.\\
\hspace*{1em}\textbf{2.} First, turn right at 2 o'clock and walk two to four steps, where you will reach the crosswalk.\\
\hspace*{1em}\textbf{3.} There seems to be no alarm button on the signal pole, so you should cross the street with extra caution. There is a pedestrian across the street, so you can shout for assistance.\\
\hspace*{1em}\textbf{4.} Walk four to five steps to cross the street. Walk 2 to 3 steps further and turn left to locate the subway exit. You have to walk down the stairs to enter the subway station.

\tcblower

\textbf{Request:} Enter the store named xxx.\vspace{1ex}

\textbf{Response:}\\
\hspace*{1em}\textbf{1.} The store named xxx is located in the 4 o'clock direction, approximately 10 meters away.\\
\hspace*{1em}\textbf{2.} First, turn right at a 20-degree angle.\\
\hspace*{1em}\textbf{3.} Then, walk 20 to 30 steps to reach the store entrance.\\
\hspace*{1em}\textbf{4.} Bollards and cars are in front of you, so please be careful.\vspace{1ex}

\textbf{Enhanced response:}\\
\hspace*{1em}\textbf{1.} The store named xxx is located in the 2 o'clock direction, approximately 20 meters away. To approach the store, you must cross the street using the crosswalk located at approximately 12 o'clock, approximately 5 meters away.\\
\hspace*{1em}\textbf{2.} First, go straight for eight to ten steps and stop. Please stop and wait for the traffic signal if you encounter bollards and tactile paving in front of the crosswalk.\\
\hspace*{1em}\textbf{3.} Look for the signal pole on the left side of the crosswalk and check for an alarm button. If there is no alarm button, please be aware that a car is coming from the right side of the road. Once you hear the `go' sound from the alarm or no sound of car engines on the right side, carefully cross the crosswalk for 10 to 15 steps.\\
\hspace*{1em}\textbf{4.} Then, turn right in the 3 o'clock direction and walk 15 to 20 steps. There are multiple stores on your left as you proceed.\\
\hspace*{1em}\textbf{5.} The xxx store is located on the next block, so be cautious while crossing the road for 5 to 10 steps. The store will be on your left in the direction of 11 o'clock. 
\end{tcolorbox}
\vspace{-0.5em}
\caption{\textbf{Few-shot examples for outdoor response refinement by GPT-4o mini.} These examples demonstrate how the model is instructed to improve a baseline \emph{Response} into a highly detailed, safety-oriented \emph{Enhanced response}. The enhanced versions integrate specific environmental hazards and precise navigational metrics tailored for BLV users (Note: xxx represents store names or signs detected in the original image).}
\vspace{-1em}
\label{app:fewshot_prompts_gpt_outdoors}
\end{table*}
\begin{table*}[ht]
\centering
\begin{tcolorbox}[
    enhanced,
    colback=black!2, 
    colframe=black!70, 
    colbacktitle=black!10, 
    coltitle=black, 
    fonttitle=\bfseries\small, 
    title=Few-Shot Examples for Refining LVLM Responses Using GPT-4o mini (Indoor Scenes),
    boxsep=4pt, 
    left=6pt, 
    right=6pt, 
    top=4pt, 
    bottom=4pt, 
    arc=3pt, 
    segmentation style={solid, draw=black!30, line width=0.5pt}, 
    fontupper=\small, 
    fontlower=\small
]
\textbf{Request:} Open the refrigerator door in the kitchen.\vspace{1ex}

\textbf{Response:}\\
\hspace*{1em}\textbf{1.} The refrigerator door is located in the 8 o'clock direction.\\
\hspace*{1em}\textbf{2.} Turn right from where you are and walk two to three steps.\\
\hspace*{1em}\textbf{3.} Then, turn left in the 2 o'clock direction and continue walking until you reach the fridge.\\
\hspace*{1em}\textbf{4.} You can access the items inside by opening the fridge door. Watch out for any obstacles in your way.\vspace{1ex}

\textbf{Enhanced response:}\\
\hspace*{1em}\textbf{1.} The kitchen is directly in front of you, and the refrigerator door is located in the 10 o'clock direction, approximately 2 meters away.\\
\hspace*{1em}\textbf{2.} Walk straight for two to three steps and then turn left in the 9 o'clock direction. Don't be confused with the cabinet to the left of the refrigerator.\\
\hspace*{1em}\textbf{3.} Reach out your arm to reach the door handle on the left side of the fridge.\\
\hspace*{1em}\textbf{4.} The kitchen space appears to be not too large, so be careful when opening the refrigerator door.

\tcblower

\textbf{Request:} Pick up the bottle of water on the box.\vspace{1ex}

\textbf{Response:}\\
\hspace*{1em}\textbf{1.} The bottle of water is on the box right in front of you, between the 11 to 12 o'clock direction, approximately 10 meters away. There is a machine at knee height between you and the box, so be cautious.\\
\hspace*{1em}\textbf{2.} Walk one to two steps and stretch your hand in the 11 o'clock direction.\\
\hspace*{1em}\textbf{3.} You should be able to grab the bottle of water beside the styrofoam box. There are many obstacles around you, so proceed carefully.\vspace{1ex}

\textbf{Enhanced response:}\\
\hspace*{1em}\textbf{1.} The bottle of water is on top of the shelf and box, next to the styrofoam. It is located in the 11 o'clock direction, approximately 3 meters away. Be aware that there are multiple obstacles around you, and the space appears tight, so proceed cautiously.\\
\hspace*{1em}\textbf{2.} Walk one to two steps and stretch your hand in the 11 o'clock direction. Be careful since there is a machine at knee height between you and the space where the water bottle is located.\\
\hspace*{1em}\textbf{3.} You should be able to grab the bottle of water on the right side of the styrofoam box. Additionally, note there is a chair on top of the table to your right and a pile of clothes on your left, so navigate carefully.

\tcblower

\textbf{Request:} Enter the bathroom.\vspace{1ex}

\textbf{Response:}\\
\hspace*{1em}\textbf{1.} The bathroom is in the 12 o'clock direction, approximately 3 to 4 meters away. There are toys and other obstacles around you, so be cautious.\\
\hspace*{1em}\textbf{2.} First, turn slightly right in the 2 o'clock direction and walk 2 to 3 steps.\\
\hspace*{1em}\textbf{3.} Then, turn left in the 1 o'clock direction and walk 2 to 3 steps. Be mindful of the thin iron shelf on the right of the bathroom door. The bathroom is clean, so you don't have to clean.\vspace{1ex}

\textbf{Enhanced response:}\\
\hspace*{1em}\textbf{1.} The bathroom is located in the 12 o'clock direction, approximately 3 meters away. Please proceed carefully since there are toys on your left in the 11 to 12 o'clock direction at less than a 1-meter distance and a bookshelf on your right in the 2 o'clock direction at a 1-meter distance.\\
\hspace*{1em}\textbf{2.} First, turn right in the 2 o'clock direction and walk 3 to 4 steps. The bookshelf should be directly in front of you at this point.\\
\hspace*{1em}\textbf{3.} Then, turn in the 3 o'clock direction and walk 4 to 5 steps. There is a shelf on the right side of the bathroom door below the bathroom switch button.
\end{tcolorbox}
\vspace{-0.5em}
\caption{\textbf{Few-shot examples for indoor response refinement by GPT-4o mini.} These examples demonstrate how the model is instructed to improve baseline responses for indoor mobility tasks. The enhanced versions provide much tighter distance estimations, explicit warnings about specific clutter, and remove unhelpful assumptions.}
\label{app:fewshot_prompts_gpt_indoors}
\end{table*}
\begin{table*}[ht]
\centering
\begin{tcolorbox}[
    enhanced,
    colback=black!2, 
    colframe=black!70, 
    colbacktitle=black!10, 
    coltitle=black, 
    fonttitle=\bfseries\small, 
    title=Generated Deep Context Samples (\emph{Before} and \emph{After}),
    boxsep=4pt, 
    left=6pt, 
    right=6pt, 
    top=4pt, 
    bottom=4pt, 
    arc=3pt, 
    segmentation style={solid, draw=black!30, line width=0.5pt}, 
    fontupper=\small, 
    fontlower=\small
]
\textbf{Request:} Enter the first building to the left.\vspace{1ex}

\textbf{Response (\emph{before}):}\\
The parked car is located at the end of the alley, in the 12 o'clock direction. There are multiple cars and a building on your left side. Walk approximately five to seven steps straight towards the end of the alley. Turn right slightly in the 9 o'clock direction and continue walking for about ten steps, being cautious of pedestrians and other obstacles like parked vehicles. Once you reach the car, carefully navigate around it to proceed further down the alley. 

\tcblower

\textbf{Response (\emph{after}):}\\
\hspace*{1em}\textbf{1) Scene Description:} You are walking near a building to the left, which is at 11 o'clock and about 15--20 meters away.\\
\hspace*{1em}\textbf{2) Main Obstacles:} There is a parked car on the left side of your path, so be cautious as you walk past it.\\
\hspace*{1em}\textbf{3) Step-by-Step Actions:}\\
\hspace*{2em}-- Step 1: Turn right in the 10 o'clock direction.\\
\hspace*{2em}-- Step 2: Walk straight for 5--7 steps, avoiding the parked car on your left.\\
\hspace*{2em}-- Step 3: Continue walking for another 3 steps until you reach the entrance of the first building to the left.
\end{tcolorbox}
\vspace{-0.5em}
\caption{\textbf{Qualitative comparison of generated deep context responses.} This example illustrates the difference between an unstructured, baseline response (top) and the refined, highly structured format (bottom). Based on a human-verified request, the \emph{after} response clearly delineates the scene layout, key obstacles, and step-by-step navigation, providing a much safer and more actionable guide for BLV users.}
\label{app:qual_deepcontext_samples}
\end{table*}

\begin{table*}[ht]
\centering
\resizebox{\textwidth}{!}{%
\begin{tabular}{cccccccccc}
\toprule
\textbf{ID} & \textbf{Gender} & \textbf{Age} & \textbf{Vision Level} & \textbf{Onset Age} & \textbf{Familiar AI Tools} & \textbf{AI Usage Frequency} & \textbf{Primary Navigation Aids} & \textbf{Aid Usage Frequency} & \textbf{Desired Future Assistance} \\
\midrule
P1 & M & 51 & Severe & 7 & Seeing AI & As needed & White cane & Always & Headphone-based \\   
P2 & M & 51 & Total Blindness & 14 & Seeing AI & Twice weekly & White cane & Always & Smartphone-based \\  
P3 & M & 48 & Severe & 9 & ChatGPT, Bard & Occasionally & Audible pedestrian signals & As needed & Smartphone-based \\  
P4 & M & 52 & Total Blindness & 13 & ChatGPT & For teaching only & \begin{tabular}[c]{@{}c@{}}White cane (past) \\ Assistance from others (current)\end{tabular} & As needed & Smartphone-based \\  
P5 & F & 55 & Moderate & 20 & Seeing AI, Sullivan+ & 2--3 times weekly & \begin{tabular}[c]{@{}c@{}}White cane, \\ Assistance from others\end{tabular} & Always & Smart glasses \\   
P6 & M & 54 & Total Blindness & 16 & \begin{tabular}[c]{@{}c@{}}Seeing AI, Sullivan+, \\ ChatGPT, Gemini\end{tabular} & 1--2 times weekly & White cane & Always & AI-integrated \\  
P7 & M & 51 & Severe & 0 & Seeing AI, Sullivan+ & 3 hours weekly & White cane & Always & AI-integrated \\  
P8 & M & 21 & Severe & 0 & \begin{tabular}[c]{@{}c@{}}Seeing AI, \\ ChatGPT, Gemini\end{tabular} & 3 times weekly & \begin{tabular}[c]{@{}c@{}}White cane, \\ Google Maps, BlindSquare\end{tabular} & As needed & AI-integrated \\
\bottomrule
\end{tabular}
}
\caption{\textbf{Demographics and technology usage profiles of BLV participants.} Overall, the cohort predominantly consists of middle-aged individuals with severe to total vision loss who actively rely on the white cane for navigation and exhibit strong familiarity with mobile AI accessibility applications. Participants P1--P6 took part in the first round of experiments, while P1, P2, P7, and P8 participated in the second round.}
\label{tab:personal_blv}
\end{table*}

\begin{table*}[ht]
\centering
\begin{tcolorbox}[
    enhanced,
    colback=black!2, 
    colframe=black!70, 
    colbacktitle=black!10, 
    coltitle=black, 
    fonttitle=\bfseries\small, 
    title=Prompts for Refining LVLM Responses,
    boxsep=4pt, 
    left=6pt, 
    right=6pt, 
    top=4pt, 
    bottom=4pt, 
    arc=3pt, 
    segmentation style={solid, draw=black!30, line width=0.5pt}, 
    fontupper=\small, 
    fontlower=\small
]
You are an expert at providing a Blind or Low Vision (BLV) an accurate, helpful description, given an environmental scene (outdoor or indoor) and corresponding to their text-based request. Remember that BLV users cannot see as much as normally-sighted humans, so you must provide detailed but precise information from the image. Enhance the model response by including precise clock directions (options: 9, 10, 11, 12, 1, 2, 3 o'clock), depth levels (in meters or steps), and objects BLV users should avoid or utilize. 
\end{tcolorbox}
\vspace{-0.5em}
\caption{\textbf{System prompt for refining LVLM responses.} The prompt guides the model to incorporate perspectives from both normally sighted individuals and BLV users, producing spatially precise descriptions while adhering to safety considerations.}
\label{app:system_prompts_ref}
\end{table*}
\begin{table*}[ht]
\centering
\begin{tcolorbox}[
    enhanced,
    colback=black!2, 
    colframe=black!70, 
    colbacktitle=black!10, 
    coltitle=black, 
    fonttitle=\bfseries\small, 
    title=Prompts for Evaluating LVLM Responses,
    boxsep=4pt, 
    left=6pt, 
    right=6pt, 
    top=4pt, 
    bottom=4pt, 
    arc=3pt, 
    segmentation style={solid, draw=black!30, line width=0.5pt}, 
    fontupper=\small, 
    fontlower=\small
]
You will be given one sentence of visual caption generated from one image and request. Your task is to rate the generated caption on one metric.\vspace{1.5ex}

\textbf{Evaluation Criteria:} Score is from 0 to 100 - The generated caption should accurately fulfill the request based on the image. You should penalize captions that include irrelevant details, omit significant elements indicated by the request, or fail to accurately describe the visual content of the image. Assign an integer score from 0 to 100 for the caption based on the following dimensions:\vspace{1.5ex}

\hspace*{1em}\textbullet~ \textbf{Direction Accuracy:} Are the directions provided (\eg, angle, clock direction, left or right) in the caption accurate and aligned with the image?\vspace{1ex}

\hspace*{1em}\textbullet~ \textbf{Depth Accuracy:} Are the depth levels (\eg, x meters, x steps) in the caption correct and aligned with the image?\vspace{1ex}

\hspace*{1em}\textbullet~ \textbf{Response Quality:} Is the caption free from unnecessary repetition, illogical order, irrelevance to the scenario, and hallucinations?\vspace{1ex}

\hspace*{1em}\textbullet~ \textbf{Safety and Actionability:} Are all the action verbs included in the caption entirely safe and actionable for blind users to fulfill their request?\vspace{1ex}

\hspace*{1em}\textbullet~ \textbf{Sufficiency:} Does the caption provide all the necessary information for blind users to fulfill their request?\vspace{1ex}

\hspace*{1em}\textbullet~ \textbf{Conciseness:} Is the caption concise and free from verbosity?\vspace{1ex}

\hspace*{1em}\textbullet~ \textbf{Overall:} How would you rate the caption overall?\vspace{1.5ex}

\textbf{Request:} [request]\vspace{1.5ex}

\textbf{Generated Caption:} [response]\vspace{1.5ex}

Provide a JSON output with integer scores for the 7 evaluation criteria.
\end{tcolorbox}
\vspace{-0.5em}
\caption{\textbf{System prompt for score generation in constructing \oursdata training dataset.} GPT-4o mini is instructed to evaluate responses produced by 7B-scale models across seven aspects, following criteria aligned with those provided to human annotators.}
\vspace{-1em}
\label{app:system_prompts_scoring}
\end{table*}
\begin{table*}[ht]
\centering
\begin{tcolorbox}[
    enhanced,
    colback=black!2, 
    colframe=black!70, 
    colbacktitle=black!10, 
    coltitle=black, 
    fonttitle=\bfseries\small, 
    title=Human Labeling Guidelines for Evaluating LVLM Responses,
    boxsep=4pt, 
    left=6pt, 
    right=6pt, 
    top=4pt, 
    bottom=4pt, 
    arc=3pt, 
    segmentation style={solid, draw=black!30, line width=0.5pt}, 
    fontupper=\small, 
    fontlower=\small
]
\textbf{Overview}: This study aims to evaluate the capability of vision-language models in generating deep context to support the mobility of blind or low-vision (BLV) users. As an annotator, your task is to assess and refine the model-generated responses based on the given indoor or outdoor images and scenarios. You will evaluate the provided responses on several criteria and make necessary corrections to ensure accuracy, usability, and relevance. Each scenario consists of the following:\vspace{1ex}

\hspace*{1em}\textbf{1. Image:} an indoor or outdoor environmental scene provided as visual context.\\
\hspace*{1em}\textbf{2. Request:} a BLV mobility-related request or task from BLV users.\\
\hspace*{1em}\textbf{3. Step-by-step description:} the vision-language model's response to the request.\vspace{1.5ex}

\textbf{Notes for Refinement}: The deep context responses often follow this format:\vspace{1ex}

\hspace*{1em}\textbf{1. Scene Description:} An overview of the environment, highlighting key landmarks.\\
\hspace*{1em}\textbf{2. Distance and Direction to the Goal:} Clear directional and distance information to the target.\\
\hspace*{1em}\textbf{3. Obstacles to Watch For:} Specific obstacles the user should be aware of.\\
\hspace*{1em}\textbf{4. Step-by-Step Directions:} Detailed instructions for completing the task.\vspace{1ex}

When a response does not follow this format, refine it accordingly. Copy the original response and correct errors, remove unnecessary details, or add missing information.\vspace{1.5ex}

\textbf{Evaluation Criteria}: For each response, you will rate the following aspects on a Likert scale (1 to 5) or a binary scale and provide corrections where necessary:\vspace{1.5ex}

\textbf{1. Direction Accuracy}\\
\hspace*{1em}\emph{Definition:} Are the directions provided (\eg, angle, clock direction, left or right) in the response accurate and aligned with the image?\\
\hspace*{1em}\emph{Ratings:} 1: no accurate info at all, 2: 3 inaccurate info, 3: 2 inaccurate info, 4: 1 inaccurate info, 5: entirely accurate\vspace{1ex}

\textbf{2. Depth Accuracy}\\
\hspace*{1em}\emph{Definition:} Are the depth levels (\eg, x meters, x steps) in the response correct and aligned with the image?\\
\hspace*{1em}\emph{Ratings:} 1: no accurate info at all, 2: 3 inaccurate info, 3: 2 inaccurate info, 4: 1 inaccurate info, 5: entirely accurate\vspace{1ex}

\textbf{3. Response Quality} (Step-by-Step Order, Relevance, Hallucination and Repeatedness)\\
\hspace*{1em}\emph{Definition:} Is the response free from unnecessary repetition, illogical order, irrelevance to the scenario, and hallucinations?\\
\hspace*{1em}\emph{Ratings:} 0: no (the response is illogically ordered, and includes irrelevant or hallucinated details, or has repetitions), 1: yes\vspace{1ex}

\textbf{4. Safety and Actionability}\\
\hspace*{1em}\emph{Definition:} Are the actions in the response safe and actionable for BLV users?\\
\hspace*{1em}\emph{Ratings:} 1: no safe or actionable actions, 2: no safe actions but actionable actions, 3: safe actions but include non-actionable actions (\eg, watch xxx, see xxx, observe xxx, drive xxx, etc.), 4: includes consensual safety actions (\eg, watch out for cars when crossing the street.), 5: entirely safe and actionable\vspace{1ex}

\textbf{5. Sufficiency}\\
\hspace*{1em}\emph{Definition:} Does the response provide all the necessary information for BLV users to complete the task?\\
\hspace*{1em}\emph{Ratings:} 1: strongly disagree, 2: disagree, 3: neutral, 4: agree, 5: strongly agree\vspace{1ex}

\textbf{6. Conciseness}\\
\hspace*{1em}\emph{Definition:} Is the response concise and free from verbosity?\\
\hspace*{1em}\emph{Ratings:} 1: strongly disagree, 2: disagree, 3: neutral, 4: agree, 5: strongly agree\vspace{1ex}

\textbf{7. Overall}\\
\hspace*{1em}\emph{Definition:} Rate the response overall.\\
\hspace*{1em}\emph{Ratings:} 1: not useful at all, 2: needs significant improvement, 3: needs some improvement, 4: valid for BLV users, 5: very useful for BLV users.
\end{tcolorbox}
\vspace{-0.5em}
\caption{\textbf{Human experiment guideline for LVLM response evaluation.} The guideline outlines both the overall workflow and the detailed steps for the annotation and refinement processes.}
\label{app:guideline}
\end{table*}





\clearpage
\newpage
\section{Method Implementation Details}\label{app:model}

\begin{table*}[ht]
\centering
\resizebox{0.7\textwidth}{!}{%
\begin{tabular}{cccccc}
\toprule
\textbf{Evaluator} & \textbf{Batch size} & \textbf{Learning rate} & \textbf{Fix rate}   & \textbf{Polaris} & \textbf{ImgREW} \\
\cmidrule(lr){1-1}\cmidrule(lr){2-4}\cmidrule(lr){5-6}
\multirow{4}{*}{ImgTxtREW-S} & 32     & 1e-5          & 0.7      & \textbf{97.8}             & 62.1            \\
 & 32     & 5e-6          & 0.7      & 92.3                  & \underline{64.8}             \\
 & 64    & 1e-5          & 0.7      & \underline{93.0}              & 64.6            \\
 & 64    & 1e-6          & 0.8      &  75.9                & \textbf{65.0}           \\
\bottomrule
\end{tabular}
}
\caption{\textbf{Hyperparameter tuning results.} We evaluate various combinations of batch size, learning rate, and fix rate to optimize model performance. The best and second-best results are highlighted in \textbf{bold} and \underline{underlined}, respectively. The optimal configuration varies depending on the target evaluation metric, illustrating a clear trade-off between Polaris (text-to-image) and ImgREW (image-to-text) performance.}
\label{app:hyper}
\end{table*}

\begin{table}[ht]
\centering
\resizebox{\textwidth}{!}{%
\begin{tabular}{@{}lcccc@{}}
\toprule
 & \textbf{PASCAL; FOILR1; FOILR4; OID; ImgREW; P-Acc} & \textbf{FlickrExp; Polaris} & \textbf{FlickrCF} & \textbf{\oursdataB} \\
VLM-based Metric & P-Acc & $\tau_b$ & $\tau_c$ & $\tau_c$ \\
\cmidrule(lr){1-1}\cmidrule(lr){2-4}\cmidrule(lr){5-5}
BLIP-S      & 78.22 & \textbf{47.45} & \textbf{37.80} & \textbf{10.08} \\
TxtBLIP-S   & \underline{78.95} & \underline{44.10} &  \underline{35.90}   & \underline{9.40} \\
\cmidrule(lr){1-5}
ImgREW-S    & 77.68 & 43.00 &  36.20        & -3.06 \\
ImgTxtREW-S & \textbf{79.47} & 42.75 & 35.70  & -2.96 \\
\bottomrule
\end{tabular}
}
\vspace{-0.5em}
\caption{\textbf{Aggregated performance of BLIP-based evaluation metrics.} BLIP-S overall shows the highest correlations with human judgments on both general and BLV-user tasks. Best results are in \textbf{bold}; second-best are \underline{underlined}.}
\label{tab:metric_results_app}
\end{table}

\paragraph{Training procedure for VLM.}
Before introducing our LVLM-based evaluator (\oursmetric), we first explore a straightforward approach by fine-tuning a lightweight and efficient VLM encoder. Specifically, we build upon BLIP-S~\citep{li2022blip} and its reward-based variant ImgREW-S~\citep{xu2024imagereward}, which are originally trained on image-to-text and text-to-image data, respectively. From these, we derive TxtBLIP-S and ImgTxtREW-S through additional fine-tuning. Consistent with the \emph{stage} 1 training procedure of \oursmetric, we fine-tune BLIP-S and ImgREW-S using the Polaris dataset~\citep{wada2024polos}. This dataset provides human-annotated scores in the range [0, 1] (0.00, 0.25, 0.50, 0.75, 1.00) for image–text pairs, along with five reference captions per image. We apply a threshold of 0.5 to construct positive and negative pairs, resulting in 22,803 training, 30,461 validation, and 38,076 test samples.
ImgREW-S is optimized using a pairwise ranking objective over positive and negative texts for each image:

\vspace{-1em}
\begin{equation}
\label{eq:1}
\mathcal{L}(\theta) = -\mathbb{E}_{(I, t_i, t_j) \sim \mathcal{D}} \left[ \log \left( \sigma \left( f_{\theta}(I, t_i) - f_{\theta}(I, t_j) \right) \right) \right]
\end{equation}
\vspace{-1em}

Here, $I$, $t$, $\mathcal{D}$, and $f_{\theta}$ denote the image, positive/negative text, data distribution, and preference model, respectively. The objective encourages the model to assign higher rewards to image-aligned (positive) captions than to misaligned (negative) ones. We use the following hyperparameters: 1 epoch, batch size of 64, gradient accumulation steps of 4, learning rate of $1\mathrm{e}{-5}$ with cosine decay, and a parameter freezing ratio of 0.7.

The effect of batch size, learning rate, and freezing ratio for TxtBLIP-S and ImgTxtREW-S is analyzed in Tab.~\ref{app:hyper}. Due to a clear trade-off between performance on the Polaris and ImgREW datasets for ImgTxtREW-S, we adopt a balanced configuration (batch size 64, learning rate $1\mathrm{e}{-5}$, freeze rate 0.7). Final results are reported in Tab.~\ref{tab:metric_results_app}.
As BLIP-S demonstrates the strongest overall performance across both general and BLV-specific benchmarks, we further fine-tune it on the 11.2k training (1.4k validation) set of unstructured-structured GPT-4o mini pairs, resulting in \oursmetricVLM. As shown in Tab.~\ref{tab:metric_results}, \oursmetricVLM improves performance on \oursdataB while maintaining alignment with human judgments on general datasets. However, due to the limited scale of BLV-specific data in \oursdataB, this approach alone remains insufficient for capturing the full complexity of the task. This motivates our transition to LVLM-based evaluator (\oursmetric), which can better leverage language understanding to interpret nuanced requirements of human users, including BLV individuals.

\begin{table*}[ht]
\centering
\resizebox{\textwidth}{!}{%
\begin{tabular}{@{}lcccccccccc@{}}
\toprule

& \textbf{PASCAL} & \textbf{FOILR1} & \textbf{FOILR4} & \textbf{Polaris} & \textbf{OID} & \textbf{ImgREW} & \textbf{FlickrExp} & \textbf{FlickrCF} & \textbf{Polaris} & \cellcolor[HTML]{f5f2ee}\textbf{\oursdataB} \\ 

VLM-based Metric & \multicolumn{6}{c}{P-Acc} & $\tau_{c}$ & $\tau_{b}$ & $\tau_{c}$ & \cellcolor[HTML]{f5f2ee}$\tau_{c}$ \\ 
\cmidrule(lr){1-1}\cmidrule(lr){2-7}\cmidrule(lr){8-11}

 CLIP-S & 80.7 & 87.2 & 87.2 & 79.7 & 56.5 & 56.7  & 51.2 & 34.4 & 52.3 & \cellcolor[HTML]{f5f2ee}-2.30 \\
 
 LongCLIP-S & 82.8 & 91.6 & 91.6 & 77.5 & 58.1 & 56.5 & 54.1 & 35.4 & 54.0 & \cellcolor[HTML]{f5f2ee}2.73 \\
 
 PAC-S & 82.4 & 93.7 & 94.9 & 77.0 & 57.7 & 57.2 & \underline{55.9} & \underline{37.6} & 52.5 & \cellcolor[HTML]{f5f2ee}-0.75 \\
 
 Ref-free Polos & 81.0 & 88.7 & 88.7 &  60.0 & \textbf{66.2} & 56.6 & 51.4 & 34.4 & 52.3  & \cellcolor[HTML]{f5f2ee} -3.18 \\
 
 CA-CLIPS-S & -  & -  & -  & -  & -  & - & - & - & - & \cellcolor[HTML]{f5f2ee} -5.08 \\
 
 RefCLIP-S$\dag$ & 83.1 & 91.0 & 92.6 & - & -  & - & 53.0 & 36.4 & 52.3 & \cellcolor[HTML]{f5f2ee} - \\
 
 RefPAC-S$\dag$ & \underline{84.7} & 88.7 & 94.9 & - & -  & -  & \underline{55.9} & \underline{37.6} & \underline{56.0} &  \cellcolor[HTML]{f5f2ee} - \\
 
 Polos$\dag$ & \textbf{86.5} & 93.3 & 95.4 & - & -  & - & 56.4 & \textbf{37.8} & \textbf{57.8} &  \cellcolor[HTML]{f5f2ee} - \\
 
 BLIP-S & 82.5 & \textbf{95.1} & \textbf{95.1} & 79.5 & \underline{59.3} & 57.8 & \textbf{57.1} & \textbf{37.8} & 54.0  & \cellcolor[HTML]{f5f2ee} \underline{10.08} \\
 
 ImgREW-S & 81.5 & 93.8 & 93.8 & 73.3 & 58.5 & \textbf{65.2} & 49.8 & 36.2 & 52.3  & \cellcolor[HTML]{f5f2ee} -3.06  \\

 \cellcolor[HTML]{f5f2ee} \oursmetricVLM &  \cellcolor[HTML]{f5f2ee} 82.3 &  \cellcolor[HTML]{f5f2ee} \underline{95.0} &  \cellcolor[HTML]{f5f2ee} \underline{95.0} & \cellcolor[HTML]{f5f2ee} 79.4 & \cellcolor[HTML]{f5f2ee} \underline{59.3}  &  \cellcolor[HTML]{f5f2ee} 57.8 & \cellcolor[HTML]{f5f2ee} 51.7  & \cellcolor[HTML]{f5f2ee} 35.8  &  \cellcolor[HTML]{f5f2ee} 53.9   & \cellcolor[HTML]{f5f2ee} \textbf{10.28} \\
\bottomrule
\end{tabular}
}
\vspace{-0.5em}
\caption{\textbf{Performance of VLM-based evaluation metrics across multimodal preference datasets.} While existing automatic metrics demonstrate strong correlations with human judgments on general image-text pairs, they fail to align with Blind and Low Vision (BLV) user preferences in our \oursdataB dataset. Notably, our proposed \oursmetricVLM achieves the highest correlation on this BLV-aware data. \textbf{Note:} Reference-based metrics ($\dag$) cannot be evaluated on datasets lacking references (``-''). CA-CLIP-S requires contextual input (e.g., user requests) and is only evaluated where applicable. Performance is measured via preference accuracy (P-Acc) and Kendall rank correlation coefficients ($\tau_c$, $\tau_b$). Best results are in \textbf{bold}; second-best are \underline{underlined}.}
\label{tab:metric_results}
\end{table*}

\paragraph{Evaluation stage for VLM.}
Non-reward-based models, such as CLIP-S~\citep{hessel2021clipscore} and BLIP-S~\citep{li2022blip}, compute similarity as the cosine similarity between normalized image and text embeddings. In contrast, reward-based metrics—including ImgREW-S~\citep{xu2024imagereward}, ImgTxtREW-S, and \oursmetric—produce a scalar reward derived from text features encoded by the final multilayer perceptron. The final score is obtained by normalizing this reward using fixed mean and standard deviation values.

\paragraph{Evaluation stage for LVLM.}
We provide all evaluation prompts for generative LVLMs, as these models require detailed task-specific instructions. Tab.~\ref{app:prompt_gen1} presents prompts used with Molmo-7B~\citep{deitke2024molmo}, Qwen2-VL-7B~\citep{Qwen2VL}, and InternVL2-8B~\citep{chen2024internvl} for both pairwise (top) and pointwise (bottom) ranking settings. Tables~\ref{app:prompt_gen2} and \ref{app:prompt_gen3} provide prompts for LLaVA-Critic-7B~\citep{xiong2024llava} and LLaMA-3.2-11B~\citep{llama3_2}, respectively. Finally, we include two prompt variations for IXCREW-S~\citep{zang2025internlm} in Tab.~\ref{app:prompt_gen4}.

\begin{table*}[ht]
\centering
\begin{tcolorbox}[
    colback=black!2, 
    colframe=black!70, 
    colbacktitle=black!10, 
    coltitle=black, 
    fonttitle=\bfseries\small, 
    title=Prompts for Evaluating Generative Models - Molmo-7B; Qwen2-VL-7B; InternVL2-8B,
    boxsep=4pt, 
    left=6pt, 
    right=6pt, 
    top=4pt, 
    bottom=4pt, 
    arc=3pt, 
    segmentation style={solid, draw=black!30, line width=0.5pt}, 
    fontupper=\small, 
    fontlower=\small
]
You are a highly capable multimodal AI assistant tasked with evaluating answers to visual questions. Please analyze the following image and question, then determine which of the two provided answers is better.\vspace{1.5ex}

\textbf{Question:} Which caption describes the image better?\vspace{1ex}

\textbf{Answer 1:} [reference or candidate caption]\vspace{1ex}

\textbf{Answer 2:} [reference or candidate caption]\vspace{1.5ex}

Please evaluate both answers based on the following criteria:\\
1. Accuracy: How well does the answer align with the visual information in the image?\\
2. Completeness: Does the answer fully address all aspects of the question?\\
3. Clarity: Is the answer easy to understand and well-articulated?\\
4. Relevance: Does the answer directly relate to the question and the image?\vspace{1.5ex}

After your evaluation, please:\\
1. Explain your reasoning for each criterion.\\
2. Provide an overall judgment on which answer is better (Answer 1 or Answer 2). For example: Overall Judgment: Answer X is better.\vspace{1.5ex}

Your response should be structured and detailed, demonstrating your understanding of both the visual and textual elements of the task.

\tcblower

You are a highly capable multimodal AI assistant tasked with evaluating the quality of a caption to the image. Please analyze the following image and caption, then determine the score for the caption in the range of 0.0 (bad quality) to 1.0 (good quality).\vspace{1.5ex}

\textbf{Caption:} [candidate caption]\vspace{1.5ex}

Please evaluate the caption based on the following criteria:\\
1. Accuracy: How well does the caption align with the visual information in the image?\\
2. Completeness: Does the caption fully address all aspects of the question?\\
3. Clarity: Is the caption easy to understand and well-articulated?\\
4. Relevance: Does the caption directly relate to the question and the image?\vspace{1.5ex}

After your evaluation, please:\\
1. Explain your reasoning for each criterion.\\
2. Provide an overall judgment score. For example: Overall Judgment: X.\vspace{1.5ex}

Your response should be structured and detailed, demonstrating your understanding of both the visual and textual elements of the task.
\end{tcolorbox}
\vspace{-0.5em}
\caption{\textbf{Prompts used for evaluation in pairwise (above) and pointwise (below) ranking settings.} In pairwise evaluation, the model selects which of two texts better aligns with the image, whereas in pointwise evaluation, it assigns a score between 0 and 1 to reflect the quality of the match. For pairwise evaluation, we retain the original prompts from VL-Reward-Bench with minimal modifications.}
\label{app:prompt_gen1}
\end{table*}
\begin{table*}[ht]
\centering
\begin{tcolorbox}[
    colback=black!2, 
    colframe=black!70, 
    colbacktitle=black!10, 
    coltitle=black, 
    fonttitle=\bfseries\small, 
    title=Prompts for Evaluating Generative Models - LLaVA-Critic-7B,
    boxsep=4pt, 
    left=6pt, 
    right=6pt, 
    top=4pt, 
    bottom=4pt, 
    arc=3pt, 
    segmentation style={solid, draw=black!30, line width=0.5pt}, 
    fontupper=\small, 
    fontlower=\small
]
Given an image, please serve as an unbiased and fair judge to evaluate the quality of the captions provided by a Large Multimodal Model (LMM). Determine which caption is better and explain your reasoning with specific details. Your task is provided as follows:\vspace{1ex}

\textbf{The first caption:} [reference or candidate caption]\\
\textbf{The second caption:} [reference or candidate caption]\vspace{1.5ex}

\textbf{ASSISTANT:}
\tcblower
Given an image and a corresponding question, please serve as an unbiased and fair judge to evaluate the quality of answer answers provided by a Large Multimodal Model (LMM). Score the response out of 100 and explain your reasoning with specific details. Your task is provided as follows:\vspace{1ex}

\textbf{Question:} [What this image presents?]\\
\textbf{The LMM response:} [candidate caption]\vspace{1.5ex}

\textbf{ASSISTANT:}
\end{tcolorbox}
\vspace{-0.5em}
\caption{\textbf{Prompts used for evaluation in pairwise (above) and pointwise (below) ranking settings.} For \emph{LLaVA-Critic-7B}, the pointwise evaluation is configured to return scores on a 0–100 scale. Empirically, this model produces more reliable and well-calibrated outputs under the 0–100 range compared to a 0–1 scale, in contrast to other generative models.}
\label{app:prompt_gen2}
\end{table*}
\begin{table*}[ht]
\centering
\begin{tcolorbox}[
    colback=black!2, 
    colframe=black!70, 
    colbacktitle=black!10, 
    coltitle=black, 
    fonttitle=\bfseries\small, 
    title=Prompts for Evaluating Generative Model - LLaMA-3.2-11B,
    boxsep=4pt, 
    left=6pt, 
    right=6pt, 
    top=4pt, 
    bottom=4pt, 
    arc=3pt, 
    segmentation style={solid, draw=black!30, line width=0.5pt}, 
    fontupper=\small, 
    fontlower=\small
]
Select which of the captions describes the image better.\vspace{1ex}

\textbf{Caption 1:} [reference or candidate caption].\\
\textbf{Caption 2:} [reference or candidate caption].\vspace{1ex}

Please either only select integer 1 or 2. Do not include any text-based captions, reasons or punctuation.
\tcblower
\textbf{v1}: Rate the following caption for the given image.\vspace{1ex}

\textbf{Caption:} [candidate caption].\vspace{1ex}

Please only provide a rating in the range of 0 (poor quality) to 100 (good quality). Do not include any reasons.\vspace{3ex}

\textbf{v2}: Rate the following caption for the given image in terms of how much the caption accurately depicts the image.\vspace{1ex}

\textbf{Caption:} [candidate caption].\vspace{1ex}

Please only provide an integer score from 0 to 100. Do not include any text-based captions, reasons, or punctuation.
\end{tcolorbox}
\vspace{-0.5em}
\caption{\textbf{Prompts used for evaluation in pairwise (above) and pointwise (below) ranking settings.} Unlike other generative models, we explicitly instruct the model to output only the final decision, excluding any explanations or additional text, due to inconsistencies in response formatting. Despite this constraint, the model achieves the highest preference accuracy among the generative baselines.}
\vspace{-1em}
\label{app:prompt_gen3}
\end{table*}
\begin{table*}[ht]
\centering
\begin{tcolorbox}[
    colback=black!2, 
    colframe=black!70, 
    colbacktitle=black!10, 
    coltitle=black, 
    fonttitle=\bfseries\small, 
    title=Prompts for Evaluating Scalar-based Model - IXCREW-S,
    boxsep=4pt, 
    left=6pt, 
    right=6pt, 
    top=4pt, 
    bottom=4pt, 
    arc=3pt, 
    segmentation style={solid, draw=black!30, line width=0.5pt}, 
    fontupper=\small, 
    fontlower=\small
]
\verb|{"role": "user",|\\
\verb|"content": `Describe the image.'}|\\
\verb|{"role": "assistant",|\\
\verb|"content": batch_response}|
\tcblower
\verb|{"role": "user",|\\
\verb|"content": `I want to generate the caption from the input image.'}|\\
\verb|{"role": "assistant",|\\
\verb|"content": batch_response}|
\end{tcolorbox}
\vspace{-0.5em}
\caption{\textbf{Two prompt variants for evaluating IXCREW-S}. We explore how different user prompt formulations affect model performance (Table~\ref{tab:app_ixc}).}
\label{app:prompt_gen4}
\end{table*}


\clearpage
\newpage
\section{Additional Results}\label{app:res}

\begin{table*}[ht]
\centering
\resizebox{\textwidth}{!}{%
\begin{tabular}{@{}lccccccccc}
\toprule
 
& \textbf{PASCAL} & \textbf{FOILR1} & \textbf{FOILR4} & \textbf{Polaris} & \textbf{OID} & \textbf{ImgREW} & \textbf{FlickrCF} & \textbf{FlickrExp} & \textbf{Polaris}  \\ 

\textbf{\oursmetric} & \multicolumn{6}{c}{P-Acc} & $\tau_b$ & \multicolumn{2}{c}{$\tau_c$} \\

\cmidrule(lr){1-1}\cmidrule(lr){2-7}\cmidrule(lr){8-10}
 
   \small{~~Qwen-2B-S} & 84.7 & 97.3 & 97.3 & 83.3 & 58.5 & 59.9 & 39.4 & 57.8 & 59.8  \\
   & \texttt{\textcolor{deepblue}{(+3.20)}} & \texttt{\textcolor{deepred}{(-0.70)}} & \texttt{\textcolor{deepred}{(-0.70)}} & \texttt{\textcolor{deepred}{(-3.70)}} & \texttt{\textcolor{deepblue}{(+5.30)}} & \texttt{\textcolor{deepblue}{(+0.90)}}
   & \texttt{\textcolor{deepblue}{(+0.50)}}
   & \texttt{\textcolor{deepblue}{(+1.00)}} & \texttt{\textcolor{deepred}{(-0.30)}} \\
   \cmidrule(lr){2-10}
    
   \small{~~Qwen-7B-S} & 83.8 & 97.0 & 97.0 & 84.7 & 57.3 & 60.3 & 39.9 & 56.3 & 61.5  \\
   & \texttt{\textcolor{deepred}{(-0.20)}} & \texttt{\textcolor{deepred}{(-0.80)}} & \texttt{\textcolor{deepred}{(-0.80)}} & \texttt{\textcolor{deepblue}{(+2.50)}} & \texttt{\textcolor{deepred}{(-0.80)}} & \texttt{\textcolor{deepred}{(-2.90)}}  & \texttt{\textcolor{deepblue}{(+1.40)}}
   & \texttt{\textcolor{deepred}{(-1.80)}}
   & \texttt{\textcolor{deepblue}{(+0.40)}} \\
   \cmidrule(lr){2-10}
   
   \small{~~InternLM-7B-S} & 82.0 & 97.1 & 97.1 & 83.7 & 58.9 & 54.3 & 38.8 & 53.1 & 57.3  \\
   & \texttt{\textcolor{deepred}{(-1.20)}} & \texttt{\textcolor{deepblue}{(+0.60)}} & \texttt{\textcolor{deepblue}{(+0.60)}} & \texttt{\textcolor{deepblue}{(+2.10)}} & \texttt{\textcolor{deepred}{(-2.10)}} & \texttt{\textcolor{deepred}{(-7.30)}}  & \texttt{\textcolor{deepred}{(-0.70)}}
   & \texttt{\textcolor{deepred}{(-6.20)}}
   & \texttt{\textcolor{deepred}{(-4.30)}} \\
   \cmidrule(lr){2-10}
   
   \small{~~LLaMA-3.2-S} & 82.7 & 96.5 & 96.5 & 81.3 & 56.5 & 63.5 & 38.3 & 56.9 & 60.9 \\
   & \texttt{\textcolor{deepred}{(-0.30)}} & \texttt{\textcolor{deepred}{(-0.40)}} & \texttt{\textcolor{deepred}{(-0.40)}} & \texttt{\textcolor{deepblue}{(+2.50)}} & \texttt{\textcolor{deepred}{(-12.20)}} & \texttt{\textcolor{deepblue}{(+1.30)}} & \texttt{\textcolor{deepblue}{(+0.30)}}
   & \texttt{\textcolor{deepblue}{(+0.10)}}
   & \texttt{\textcolor{deepblue}{(+0.20)}} \\
\bottomrule
\end{tabular}
}
\vspace{-0.5em}
\caption{\textbf{Performances of our \oursmetric trained with the empty prompt setting.} The values inside parentheses indicate the performance difference compared to the default prompt setting.}
\vspace{-1em}
\label{tab:app_prompt_ablation}
\end{table*}

\paragraph{Prompt type ablation for \oursmetric.}
Tab.~\ref{tab:app_prompt_ablation} details the performance of \oursmetric across different base models when trained without explicit instructions (\ie, using an empty prompt~\citep{llama3_2} instead of task-specific guidance like ``Describe the image.''). This experiment evaluates the flexibility and robustness of our evaluator in instruction-free scenarios. As indicated by the performance differences in the table, the overall evaluation trends remain remarkably consistent, though the exact impact varies by base model. For instance, \oursmetric (Qwen-2B-S) demonstrates strong robustness, even achieving performance gains on several datasets, such as a +3.20 increase in pairwise accuracy on PASCAL and a +5.30 increase on OID. Conversely, other models exhibit more sensitivity to the missing prompt; notably, InternLM-7B-S shows a -7.30 drop in accuracy on ImgREW, and LLaMA-3.2-S sees a significant -12.20 decrease on OID. Nevertheless, the significant performance degradation observed in several base models confirms that our default explicit prompt setting is the optimal configuration for maximizing alignment with human judgment scores.

\begin{table*}[ht]
\centering
\resizebox{0.8\textwidth}{!}{%
\begin{tabular}{@{}lcccccc}
\toprule

& \textbf{PASCAL-50S} &  \textbf{Polaris} & \textbf{OID} & \textbf{ImgREW} & \textbf{FlickrExp} & \textbf{FlickrCF} \\ 
& \multicolumn{4}{c}{P-Acc} &  $\tau_{c}$ & $\tau_{b}$ \\ 

\cmidrule(lr){1-1}\cmidrule(lr){2-5}\cmidrule(lr){6-7}

IXCREW-S-v1 & 73.9 & \textbf{89.1} & 56.9 & 53.6  & 21.6 & 25.7 \\
IXCREW-S-v2 & \textbf{76.2} & 88.7 & \textbf{58.5} & \textbf{56.3} & \textbf{25.6}  & \textbf{28.0} \\
IXCREW-S-v3 & 74.2 & 81.9 & 57.5 & 53.6 & 17.0  & 25.7 \\
\bottomrule
\end{tabular}
}
\caption{\textbf{Effect of prompts on reward model performance.} Note that version 3 (v3) results are reported in the main paper to match the prompt setting as our \oursmetric. Version 2 (v2) of the prompting strategy overall yields the best human alignment performance.}
\label{tab:app_ixc}
\end{table*}
\begin{table*}[ht]
\centering
\resizebox{0.53\textwidth}{!}{%
\begin{tabular}{@{}lccc}
\toprule
& \textbf{ImgREW} & \textbf{FlickrExp} & \textbf{FlickrCF} \\ 
& P-Acc & $\tau_{c}$ & $\tau_{b}$ \\ 

\cmidrule(lr){1-1}\cmidrule(lr){2-2}\cmidrule(lr){3-4}

LLaMA-3.2-11B-v1 & 46.0 & -7.88 & 5.49  \\
LLaMA-3.2-11B-v2  & \textbf{51.6} & \textbf{5.29} & \textbf{9.00}  \\

\bottomrule
\end{tabular}
}
\caption{\textbf{Effect of prompts on generative model performance.} Note that version 2 (v2) results are reported in the main paper, which generally shows higher performance.}
\label{tab:app_llama}
\end{table*}

\paragraph{Prompt type ablation for reward model and generative LVLM.}
Tables~\ref{tab:app_ixc} and \ref{tab:app_llama} report the performances under alternative prompting strategies for IXCREW-S~\citep{zang2025internlm} and LLaMA-3.2-11B~\citep{llama3_2}, respectively. The results indicate that the choice of prompting strategy influences the evaluation performance of both reward and generative models, with version 2 (v2) consistently yielding the highest human alignment across most datasets for both IXCREW-S and LLaMA-3.2-11B.

\paragraph{Qualitative results.}
Tables~\ref{app:sample_gen1}, \ref{app:sample_gen2}, and \ref{app:sample_gen3} provide qualitative sample responses of generative LVLMs: Molmo-7B~\citep{deitke2024molmo}, LLaVA-Critic~\citep{xiong2024llava}, and InternVL2-8B~\citep{chen2024internvl}. Since the answers include reasons for their choice or scores, we extract the final scalar ratings via post-processing: specifically, the floating-point value following ``Overall Judgment'' for Molmo-7B and InternVL2-8B, and the values after ``The better caption'' and ``Score'' for LLaVA-Critic-7B. 

\begin{table*}[ht]
\centering
\begin{tcolorbox}[
    colback=black!2, 
    colframe=black!70, 
    colbacktitle=black!10, 
    coltitle=black, 
    fonttitle=\bfseries\small, 
    title=Generated Samples Using Generative Model - Molmo-7B,
    boxsep=4pt, 
    left=6pt, 
    right=6pt, 
    top=4pt, 
    bottom=4pt, 
    arc=3pt, 
    segmentation style={solid, draw=black!30, line width=0.5pt}, 
    fontupper=\small, 
    fontlower=\small
]
\textbf{1. Accuracy:}\\
\hspace*{1em}\emph{Answer 1:} Inaccurate. The image doesn't show a blond woman in a blue shirt waiting for a ride.\\
\hspace*{1em}\emph{Answer 2:} More accurate. It correctly describes a young child in blue goggles in a pool setting.\vspace{1ex}

\textbf{2. Completeness:}\\
\hspace*{1em}\emph{Answer 1:} Incomplete. It lacks key details about the child's position or the pool setting.\\
\hspace*{1em}\emph{Answer 2:} More complete. It fully describes the child, their attire, and the surrounding environment.\vspace{1ex}

\textbf{3. Clarity:}\\
\hspace*{1em}\emph{Answer 1:} Poor. It's vague and doesn't clearly convey what's in the image.\\
\hspace*{1em}\emph{Answer 2:} Better. It's concise and easy to understand, directly describing what's visible.\vspace{1ex}

\textbf{4. Relevance:}\\
\hspace*{1em}\emph{Answer 1:} Irrelevant. It doesn't address the question about the woman in the blue shirt.\\
\hspace*{1em}\emph{Answer 2:} Relevant. It directly answers the question about the child in blue goggles.\vspace{1.5ex}

\textbf{Overall Judgment:} \emph{Answer 2 is better.}\\
It accurately describes the key elements of the image, including the child, their attire, and the pool setting, which directly answers the question about what's in the picture.

\tcblower

\emph{I'll evaluate the caption based on the given criteria:}\vspace{1.5ex}

\textbf{1. Accuracy:} The caption accurately describes the child wearing blue goggles and sitting in a float, which aligns with the visual information provided.\vspace{1ex}

\textbf{2. Completeness:} The caption fully addresses the question by mentioning both the child and the goggles/float, providing a comprehensive answer.\vspace{1ex}

\textbf{3. Clarity:} The caption is easy to understand, using simple language to convey the information clearly.\vspace{1ex}

\textbf{4. Relevance:} The caption directly relates to the question and the image, focusing on the key elements described in the visual content.\vspace{1.5ex}

\textbf{Overall Judgment:} \textbf{0.9}\\
The caption scores high in most categories, with perfect accuracy, completeness, and clarity. While it could be slightly more concise, it effectively captures the essential elements of the image. The only minor reason for a point deduction is that it could potentially be even more streamlined, but this doesn't significantly impact its overall quality.
\end{tcolorbox}
\vspace{-0.5em}
\caption{\textbf{Samples of generated responses with the first (above) and second (below) prompts in Table~\ref{app:prompt_gen1}.} Molmo-7B produces answers based on the criteria given in the prompt and provides the summarized reason for the overall judgment.}
\label{app:sample_gen1}
\end{table*}
\begin{table*}[ht]
\centering
\begin{tcolorbox}[
    colback=black!2, 
    colframe=black!70, 
    colbacktitle=black!10, 
    coltitle=black, 
    fonttitle=\bfseries\small, 
    title=Generated Samples Using Generative Model - InternVL2-8B,
    boxsep=4pt, 
    left=6pt, 
    right=6pt, 
    top=4pt, 
    bottom=4pt, 
    arc=3pt, 
    segmentation style={solid, draw=black!30, line width=0.5pt}, 
    fontupper=\small, 
    fontlower=\small
]
\emph{Evaluation:}\vspace{1ex}

\textbf{1. Accuracy:}\\
\hspace*{1em}-- \textbf{Answer 1:} This answer is accurate as it describes a city bus driving down a city street, which is clearly visible in the image.\\
\hspace*{1em}-- \textbf{Answer 2:} While the sky is indeed blue and clear in the image, the description doesn't capture the primary aspects of the scene, such as the presence of the bus and other elements like the bank and parked cars.\vspace{1ex}

\textbf{2. Completeness:}\\
\hspace*{1em}-- \textbf{Answer 1:} This answer covers more parts of the scene. It includes the bus, a critical element, and implicitly communicates the urban setting.\\
\hspace*{1em}-- \textbf{Answer 2:} This answer focuses only on the sky and doesn't account for other significant scene elements.\vspace{1ex}

\textbf{3. Clarity:}\\
\hspace*{1em}-- \textbf{Answer 1:} This answer is straightforward and easily understandable.\\
\hspace*{1em}-- \textbf{Answer 2:} It's clear but less descriptive and not as comprehensive.\vspace{1ex}

\textbf{4. Relevance:}\\
\hspace*{1em}-- \textbf{Answer 1:} Relevant as it directly mentions a city bus and the city street setting, aligning closely with the image.\\
\hspace*{1em}-- \textbf{Answer 2:} While relevant to the image, the focus on the sky doesn't capture the main elements of the scene.\vspace{1.5ex}

\textbf{Overall Judgment:} \emph{Answer 1 is better.}\\
While both answers are relevant, Answer 1 captures the primary subjects of the image (city bus driving down a street) and provides a more comprehensive description. It is more accurate and addresses more aspects of the scene, making it the better choice for the given question.

\tcblower

\emph{Evaluation:}\vspace{1ex}

\textbf{1. Accuracy:}\\
The caption generally aligns well with the visual elements of the image. It accurately captures the sinister and eerie feeling of the portal, the detailed look, and the dark tones. However, it doesn't mention the arch and pipework leading to the portal, nor does it touch on the light leak in the top part of the portal.\vspace{1ex}

\textbf{2. Completeness:}\\
The caption provides a comprehensive description of the image, capturing many details like the ultra-detailed nature, the warm interior light, and the cinematic shot. However, it misses out on describing the arch and pipework that serve as pathways into the portal.\vspace{1ex}

\textbf{3. Clarity:}\\
The caption is clear and uses descriptive language to convey the mood and tone of the image effectively. The terms like `cinematic shot', `ultra-detailed', and `photo-realistic' are used to paint a vivid picture of the scene, making it easy for the reader to grasp the image's essence.\vspace{1ex}

\textbf{4. Relevance:}\\
The caption is very relevant as it directly addresses the nature of the image and uses terminology that reflects the detailed and realistic rendering expected from tools like MCBess.\vspace{1.5ex}

\textbf{Overall Judgment:}\\
Given these assessments, the caption is extremely good. It accurately reflects the contents of the image while using vivid, relevant, and descriptive language. However, it could be improved by noting additional observations, such as the arch and pipework leading into the portal, and the top light leak in the portal.\vspace{1.5ex}

\textbf{Score:} 0.85
\end{tcolorbox}
\vspace{-0.5em}
\caption{\textbf{Samples of generated responses with the first (above) and second (below) prompts in Table~\ref{app:prompt_gen1}.} InternVL2-8B answers based on the criteria given in the prompt and also provides the summarized reason for the overall judgment.}
\label{app:sample_gen2}
\end{table*}
\begin{table*}[ht]
\centering
\begin{tcolorbox}[
    colback=black!2, 
    colframe=black!70, 
    colbacktitle=black!10, 
    coltitle=black, 
    fonttitle=\bfseries\small, 
    title=Generated Samples Using Generative Model - LLaVA-Critic-7B,
    boxsep=4pt, 
    left=6pt, 
    right=6pt, 
    top=4pt, 
    bottom=4pt, 
    arc=3pt, 
    segmentation style={solid, draw=black!30, line width=0.5pt}, 
    fontupper=\small, 
    fontlower=\small
]
The first caption, ``A city bus driving down a city street,'' is better than the second caption, ``the sky is clear and blue.'' The first caption accurately describes the main subject of the image, which is a city bus driving down a city street. This provides a clear and relevant description of the scene depicted in the image. On the other hand, the second caption, while true, does not capture the primary focus of the image and lacks context about the urban environment and the presence of the bus. Therefore, the first caption is more accurate and informative, making it the better choice.\vspace{1ex}

\textbf{The better caption:} [first].

\tcblower

The LMM response provides a detailed description of the image, highlighting several key elements such as ``deathly portal to the abyss,'' ``ultra detailed,'' ``warm interior light,'' ``cinematic shot,'' ``photorealistic,'' ``octane render,'' ``high definition,'' ``fine details,'' ``sinister tones,'' ``8 k,'' and ``mcbess mood.''\vspace{1ex}

However, the response is somewhat disjointed and lacks coherence. It mentions multiple aspects without clearly connecting them to form a cohesive description of the image. For instance, while it describes the image as having a ``deathly portal to the abyss,'' it also mentions ``warm interior light'' and ``cinematic shot,'' which are contradictory in terms of atmosphere.\vspace{1ex}

Additionally, some terms like ``8 k'' and ``mcbess mood'' are not standard descriptive terms for images and may confuse the reader. The response could be improved by providing a more unified and clear description that captures the overall mood and elements of the image effectively.\vspace{1.5ex}

\textbf{Score:} 60\vspace{1.5ex}

\textbf{Reasons:}\\
\hspace*{1em}\textbullet~ The response includes several relevant details about the image.\\
\hspace*{1em}\textbullet~ However, it is somewhat disorganized and lacks coherence.\\
\hspace*{1em}\textbullet~ Some terms are not standard or clear, which might confuse the reader.\\
\hspace*{1em}\textbullet~ Overall, the response could be more polished and focused.
\end{tcolorbox}
\vspace{-0.5em}
\caption{\textbf{Samples of generated responses with the first (above) and second (below) prompts in Table~\ref{app:prompt_gen2}.} LLaVA-Critic-7B presents multiple justifications for its selection of the preferred caption and assigns an overall score accordingly.}
\label{app:sample_gen3}
\end{table*}


\clearpage
\newpage
\section{Discussion}\label{app:lim}

While \oursmetric demonstrates strong alignment with human scores in both pairwise and pointwise ranking datasets, its performance on special domain benchmarks such as \oursdata remains limited. This might be due to a lack of reliable, human-centered training datasets, particularly for fine-grained spatial understanding in BLV user perspectives. Despite being clearly instructed, we notice LVLMs tend to understand the clockwise direction from the perspective of the image itself, rather than the user's viewpoint. For instance, when instructed to choose a direction between 9 and 3 o'clock, LVLMs occasionally provide ``8 o'clock'', which corresponds to ``10 o'clock'' from an ecocentric image viewpoint. This suggests that current LVLMs are primarily trained on image-caption datasets that describe positional relationships within the image itself, rather than recognizing depth and directional cues from a specific viewpoint. Consequently, this limitation underscores the critical need for datasets specifically designed to train models to interpret and describe scenes from an egocentric, user-defined perspective.

Furthermore, we leave the development of LVLMs capable of generating more accurate and human-aligned responses for challenging multi-objective evaluation as future work. We position \oursmetric (fine-tuned with \oursdata at \emph{stage 2}) as a robust, foundational baseline for researchers and engineers building efficient multi-objective scorers. Ultimately, we hope this work inspires continued progress toward reliable assistive AI systems, especially for Blind and Low-Vision users. While careful deployment remains necessary to mitigate risks such as bias, hallucinations, and unsafe outputs to ensure these technologies maximize societal benefit, we believe that prioritizing transparent and user-centric evaluation is a crucial step toward safe, real-world deployment.


\end{document}